\documentclass[letterpaper]{article} 
\usepackage[preprint]{aaai24}  
\usepackage{times}  
\usepackage{helvet}  
\usepackage{courier}  
\usepackage[hyphens]{url}  
\usepackage{graphicx} 
\usepackage{natbib}  
\usepackage{caption} 
\usepackage{algorithm}
\usepackage{algorithmic}

\usepackage{newfloat}
\usepackage{listings}

\usepackage{graphicx}
\usepackage{commath}
\usepackage{multirow}
\usepackage{subcaption} 
\usepackage{booktabs}
\usepackage{mathtools}
\usepackage{tablefootnote}

\DeclareCaptionStyle{ruled}{labelfont=normalfont,labelsep=colon,strut=off} 
\floatstyle{ruled}
\newfloat{listing}{tb}{lst}{}
\floatname{listing}{Listing}
\title{Differentiable Sparsification for Deep Neural Networks} 

\author {
  Yongjin Lee \\
}

\affiliations{
    ETRI\\
  Daejeon, Korea \\
  \texttt{solarone@etri.re.kr}
}

\usepackage{bibentry}

\begin{document}

\maketitle

\begin{abstract}
Deep neural networks have significantly alleviated the burden of feature engineering, 
but comparable efforts are now required to determine effective architectures for these networks. 
Furthermore, as network sizes have become excessively large, a substantial amount of resources is invested in reducing their sizes. 
These challenges can be effectively addressed through the sparsification of over-complete models. 
In this study, we propose a fully differentiable sparsification method for deep neural networks,
which can zero out unimportant parameters by directly optimizing a regularized objective function with stochastic gradient descent. 
Consequently, the proposed method can learn both the sparsified structure and weights of a network in an end-to-end manner. 
It can be directly applied to various modern deep neural networks and requires minimal modification to the training process.
To the best of our knowledge, this is the first fully differentiable sparsification method. 
\end{abstract}

\section{Introduction}

Since the ground-breaking success of deep neural networks (DNNs) in the realm of machine learning and pattern recognition~\cite{Lecun1989,Krizhevsky2012},
the focal point has shifted from feature engineering to architecture engineering~\cite{Simonyan2015,He2016,He2016ID,Xie2017}.
While DNNs have eased the need for extensive feature engineering, crafting an effective architecture now demands comparable human involvement.
This entails determining the appropriate number of neurons, layers and connections between nodes.
Furthermore, as DNNs continue to grow in size,
substantial efforts are being dedicated to reducing the size of these models.

These challenges can be effectively addressed through the sparsification of over-complete models. 
A network structure can be learned or 
optimized by removing redundant blocks~\cite{Alvarez2016,Wen2016} or unnecessary connections between blocks~\cite{Ahmed2017,Liu2019,Wortsman2019}. 
Pruning has long been a popular approach to reduce the complexity and size of deep neural networks~\cite{LeCun1989OBD,Hassibi1993,Han2015,Liu2017Slimming,Frankle2019,He2019FPGM,Sui2021CHIP,Hou2022CHEX}. 
It typically involves multiple steps applied to a pre-trained model, including selecting unimportant parameters, deleting those parameters, and retraining a pruned model.
Another widely used technique is sparsity regularization with a proximal gradient~\cite{Parikh2014,Alvarez2016,Wen2016,Yoon2017}. 
It sets redundant parameters to zero during training without relying on a pre-trained model.
However, the requirement for closed-form solutions for proximal operations limits its applicability.


In this study, we introduce a fully differentiable sparsification method for DNNs.
The proposed method can zero out unimportant parameters by directly optimizing a regularized objective function with stochastic gradient descent (SGD).
This enables the simultaneous learning of both the sparsified structure and the weights of DNNs in an end-to-end manner.
By leveraging off-the-shelf automatic differentiation tools, our implementation becomes simple and straightforward.
It eliminates the need for manual coding of a pruning step or a proximal operator. 
Furthermore, our method is versatile in that it can accommodate various norms as regularizers, 
regardless of the availability of closed-form solutions for a proximal operator. 
To the best of our knowledge, these characteristics mark the first fully differentiable sparsification method.

\section{Related Work}

Our proposed method is closely related to sparsity regularization with a proximal gradient~\cite{Parikh2014}.
In this approach, a regularized objective function is formulated as
\begin{equation}\label{eqn:obj_func}
  \mathcal{L} \left(  D, W\right) + \lambda \mathcal{R} \left(W \right),
\end{equation}
where $\mathcal{L}$ represents a prediction loss, $\mathcal{R}$ is a regularization term,
$D$ is a set of training data, $W$ is a set of model parameters,
and $\lambda$ controls the trade-off between a prediction loss and a regularization term.

The $l_{1}$-norm is the most popular regularizer:
\begin{equation}\label{eqn:l1_norm}
  \mathcal{R}\left( W\right) = \sum_{i}\abs{w_{i}},
\end{equation} where $w_{i}$ represents an individual element of $W$.
The regularization term is optimized using the proximal operator:
\begin{equation}\label{eqn:l1_prox}
w_{i} \leftarrow sign\left(w_{i}\right) \left(\abs{w_{i}} - \eta \lambda \right)_{+},
\end{equation}
where $\leftarrow$ denotes an assignment operator, $\eta$ is a learning rate,
and $\left(\cdot\right)_{+}$ represents $\max(\cdot, 0)$.
As the $l_1$-regularization acts on an individual parameter, it often leads to unstructured irregular models.

To obtain regular sparse structures, the sparse regularization with $l_{2,1}$-norm can be adopted~\cite{Alvarez2016,Wen2016},
where all parameters in the same group are either retained or zeroed-out together.
The $l_{2,1}$-regularization can be expressed as:
\begin{equation}\label{eqn:group_norm}
  \mathcal{R}\left( W\right) = \sum_{g}\norm{\textbf{w}_g}_{2} = \sum_{g}\sqrt{\sum_{i}w_{g,i}^{2} },
\end{equation}
where $W=\{\textbf{w}_g\}$ and $\textbf{w}_g$ represents a group of model parameters.
The regularization term is optimized using a proximal operator,
\begin{equation}\label{eqn:group_prox}
 w_{g,i} \leftarrow \left( \frac{\norm{\textbf{w}_g}_2 - \eta\lambda}{\norm{\textbf{w}_g}_2} \right)_{+} w_{g,i}.
\end{equation}
When a group has only one single parameter,
it degenerates to $l_1$-norm regularization.

Another related group regularization method is \emph{exclusive lasso} with $l_{1,2}$-norm~\cite{Zhou2010,Yoon2017}.
This method promotes competition or sparsity within a group, rather than either retaining or removing an entire group altogether.
The regularization term is expressed as:
\begin{equation}\label{eqn:exc_norm}
\mathcal{R}\left( W\right) = \frac{1}{2} \sum_{g}\norm{\textbf{w}_g}_{1}^{2} = \frac{1}{2} \sum_{g}\left(\sum_{i}\abs{w_{g,i}} \right)^{2},
\end{equation}
and the updating rule is derived as:
\begin{equation}\label{eqn:exc_prox}
w_{g,i} \leftarrow sign\left(w_{g,i}\right) \left(\abs{w_{g,i}} - \eta \lambda \norm{\textbf{w}_g}_{1}\right)_{+}.
\end{equation}

These proximal operators involve weight decaying and thresholding steps and are performed separately
after optimizing the prediction loss, typically at every mini-batch or epoch.
Therefore, additional effort is required to manually implement and manage the update rules.
Furthermore, the application of these methods is limited to cases where closed-form solutions for the proximal operator are known.

\section{Proposed Approach}
\subsection{Base Model}
In our proposed approach, we consider a module consisting of $n$ components.
These components can be various building blocks used in DNNs, such as channels or layers in a convolutional neural network (CNN) like ResNet~\cite{He2016,He2016ID} or DenseNet layers~\cite{Huang2017}.
They can also represent nodes in a neural graph~\cite{Wortsman2019} or a graph CNN~\cite{Kipf2017}.
A module represents a composite of these components, forming a group of channels or a layer.
For illustration purposes, we assume that a module $\textbf{y}$ can be expressed as a linear combination of components $\textbf{f}_i$:
\begin{equation}\label{eqn:base}
  \textbf{y}\left( \textbf{x}\right) = \sum_{i=1}^{n}{a_i  \textbf{f}_i\left(\textbf{x}; \textbf{w}_i\right)},
\end{equation}
where $\textbf{x}$ represents the input to the module, $\textbf{w}_i$ denotes the model parameters for component $\textbf{f}_i$,
and $a_{i}$ represents an architecture parameter. The model parameters $\textbf{w}_i$ are regular parameters used in the model,
such as filters in a convolutional layer or weights in a fully connected layer.
The value of $a_{i}$ determines the importance of component $i$ or the strength of the connection between nodes. 
Setting $a_{i}$ to zero is equivalent to removing component $\textbf{f}_i$ or zeroing out $\textbf{w}_i$.

\subsection{Differentiable Sparse Parameterization}
We propose a sparse parameterization scheme that enables $a_i$ to be zero. 
First, we illustrate the integration of non-negative constraints and the softmax function: 
\begin{equation} \label{eqn:gamma}
\gamma_{i} = \exp\left(\alpha_i\right),
\end{equation}

\begin{equation} \label{eqn:thresholding}
\tilde{\gamma}_{i} = \left( \gamma_{i} - \sigma\left( \beta \right) \cdot \norm{\gamma}_{1} \right)_{+},
\end{equation}

\begin{equation}\label{eqn:softmax}
a_{i} = \frac{\tilde{\gamma}_i}{\sum_{j=1}^{n}{\tilde{\gamma}_j}},
\end{equation}
where $\alpha_i$ and $\beta$ are unconstrained free scalar parameters, $\sigma(\cdot)$ denotes the sigmoid function, and $\left(\cdot\right)_{+}$ represents the $relu(\cdot) = \max(\cdot, 0)$ function.

This parameterization allows $a_i$ to take the value of zero and ensures differentiability from a modern deep learning perspective. The free parameters $\alpha_i$ and $\beta$ are real-valued and do not restrict the training process using SGD. Hence, we can train $a_i$ through $\alpha_i$ and $\beta$.
The exponential function in Eq.(\ref{eqn:gamma}) ensures that the architecture parameters are non-negative. 
While $\gamma_i$ cannot be zero due to the exponential function, $\tilde{\gamma}_{i}$ in Eq.(\ref{eqn:thresholding}) can be zero through the thresholding operation, allowing $a_i$ to be zero as well. The term $\sigma\left( \beta \right) \cdot \norm{\gamma}_{1}$ acts as a threshold. The thresholding operation can be interpreted as follows: if the strength of component $i$ within a competition group is lower than the overall strength, it is dropped from the competition. Note that the scalar parameter $\beta$ in Eq.~(\ref{eqn:thresholding}), which determines the magnitude of the threshold, is \emph{not} a hyper-parameter but is automatically determined during training. Mathematically, the thresholding operator is not differentiable, but this is not an issue in modern deep learning frameworks, considering the support of $relu$ as a built-in differentiable function. Additionally, the $l_1$-norm of $\gamma$, denoted as $\norm{\gamma}_1$, is simply the sum of the elements of $\gamma$.

Similarly, a signed architecture parameter can be formulated as:
\begin{equation}\label{eqn:exc_param}
a_{i} = sign\left(\alpha_{i}\right) \left(\abs{\alpha_{i}} - \sigma\left(\beta \right) \norm{\alpha}_{1}\right)_{+},
\end{equation}
where $\alpha_i$ and $\beta$ are free scalar parameters as above. The gradient of the $sign$ function is zero almost everywhere, but the equation can be rewritten as:
\[
    a_{i} =
\begin{dcases}
    \left(\alpha_{i} - \sigma\left(\beta\right) \norm{\alpha}_{1}\right)_{+} & \text{if } \alpha_{i} > 0\\
    \left(\alpha_{i} + \sigma\left(\beta\right) \norm{\alpha}_{1}\right)_{-} & \text{otherwise},
\end{dcases}
\]
where $\left(\cdot\right)_{-}=\min(\cdot, 0)$. The gradient can be computed separately depending on whether the value is positive or not. 
In our understanding, modern deep learning tools, such as TensorFlow and PyTorch, already handle the $sign$ function in this manner and
we do not need to manually implement the conditional statement. 

The proximal operator of Eq.(\ref{eqn:exc_prox}) and the proposed thresholding operations of Eq.(\ref{eqn:thresholding}) and Eq.(\ref{eqn:exc_param}) share a similar form, but they have completely different meanings. The proximal operator is a learning rule used in the sparse regularization with $l_{1,2}$-norm. On the other hand, the proposed thresholding operations are the parameterized forms of the architecture parameters, integrated into neural networks, and are not necessarily bound to a specific norm.

\subsection{Sparsity Regularizer}

In our proposed approach, the regularized objective function is defined as:
\begin{equation}\label{eqn:loss}
\mathcal{L} \left( D, W, a \right) + \lambda \mathcal{R} \left( a \right),
\end{equation}
where $a$ represents a vector of architecture parameters. 
Regularizing the architecture parameter $a$ enables the learning or sparsification of a network structure.

Our proposed method is not limited to a specific norm for the regularization term and can accommodate various norms.
For instance, the $l_1$-norm (Eq.\ref{eqn:l1_norm}) can be used to remove individual components, 
while the $l_{2,1}$-norm (Eq.\ref{eqn:group_norm}) can be employed to zero-out a group of components or an entire module (see Figures~\ref{fig:channel} and ~\ref{fig:group}). 
Note that we do not need to manually implement different update rules as in the proximal gradient approach. 
Instead, we only need to rewrite a regularization term in an objective function. Example codes are provided in the supplementary material for reference.

For another example, the commonly used $l_1$-norm is unsuitable for the architecture parameters $a$ when they are normalized using the softmax function as in Eq.~(\ref{eqn:softmax}). The $l_1$-norm of the normalized architecture parameters is always $1$, i.e., $\norm{a}_1 = \sum_{i=1}^{n}{\abs{a_i}} = 1$. Therefore, it is more appropriate to employ $p$-norm with $p < 1$:
\begin{equation}\label{eqn:lp_norm}
\mathcal{R} \left(a \right) = \left(\sum_{i=1}^{n} {\abs{a_{i}}^{p}} \right) ^{\frac{1}{p}} = \left(\sum_{i=1}^{n} {a_{i}^{p}} \right) ^{\frac{1}{p}},
\end{equation}
where the second equality holds as $a_i$ is always non-negative. 
It is important to highlight that a closed-form solution for the proximal operator of a $p$-norm with $p<1$ remains elusive, rendering the proximal gradient method inapplicable. 
Nevertheless, since the regularization term is differentiable almost everywhere, our proposed approach can be directly employed.



\subsection{Rectified Gradient Flow}

\begin{table*}[!htbp]
\centering
\begin{small}
\begin{tabular}{*1c*2c*1c*2c*1c*2c*1c*2c*1c*2c}
\toprule
\multirow{2}{*}{Model}  &  \multicolumn{2}{c}{Channel Sparsity(\%)} & {} &  \multicolumn{2}{c}{Layer Sparsity(\%)} & {} &  \multicolumn{2}{c}{Top-1 Error(\%)} & {} &  \multicolumn{2}{c}{FLOPs}      & {} & \multicolumn{2}{c}{Params}  \\\cmidrule{2-3}\cmidrule{5-6}\cmidrule{8-9}\cmidrule{11-12} \cmidrule{14-15}
{}                      &  Avg.            & Std.                   & {} &  Avg.         & Std.                    & {} &  Avg.         & Std.                 & {} &  Avg.               &  Std.     & {} & Avg.                 & Std. \\
\midrule
Base                    &  00.0            & 0.0                    & {} &  0.0          & 0.0                     & {} &  5.13         & 0.05                 & {} & $ 5.0\times10^{8} $ & 0.0       & {} &  $1.7\times10^{6}$  & 0.0     \\
\midrule
\multirow{2}{*}{NS}     &  60.0            & 0.0                    & {} &  0.1          & 0.3                     & {} &  5.10         & 0.16                 & {} & $ 56.3\%$           & $ 0.2\% $ & {} &  $66.6\%$           & $0.3\%$ \\
{}                      &  75.0            & 0.0                    & {} &  2.2          & 1.1                     & {} &  6.54         & 0.18                 & {} & $ 33.4\%$           & $ 0.6\% $ & {} &  $40.3\%$           & $0.5\%$ \\
\midrule
\multirow{2}{*}{DS}     &  56.6            & 0.3                    & {} &  2.6          & 1.5                     & {} &  5.08         & 0.06                 & {} & $ 56.5\%$           & $0.8\%$   & {} &  $68.0\%$           & $0.8\%$ \\
{}                      &  74.1            & 1.1                    & {} &  32.6         & 5.4                     & {} &  5.69         & 0.10                 & {} & $ 31.8\%$           & $0.9\%$   & {} &  $44.7\%$           & $0.7\%$ \\
\bottomrule
\end{tabular}
\end{small}
\caption{Performance on CIFAR10, ResNet-164.}
\label{tab:resnet_cifar10}
\end{table*}

If $\gamma_i$ in Eq.(\ref{eqn:thresholding}) or $\alpha_i$ in Eq.(\ref{eqn:exc_param}) falls below the threshold, the gradient will be zero and the corresponding component will not receive any learning signals.
Although a component might recover if the importance scores of other components decrease further, we take steps to ensure that the architecture parameters of dropped components continue to receive learning signals. 
This is accomplished by approximating the gradient of the thresholding function using the straight-through estimator~\cite{Hinton2012,Bengio2013,Hubara2016}.
Similar to the approach in~\cite{Xiao2019}, where the gradient of a step function was approximated using the gradients of \emph{leaky relu} or \emph{soft plus}, 
we suggest using the \emph{elu}~\cite{Clevert2016} activation function to estimate the gradients of \emph{relu}. 
During the forward pass, we employ the \emph{relu} activation function, while in the backward pass, we utilize \emph{elu}.
For example, when $\alpha$ is non-negative, the gradient of Eq.(\ref{eqn:exc_param}) is approximated as:
\begin{equation}\label{eqn:rgf}
\frac{\partial a_i}{\partial \alpha_i} \approx  \frac{\partial elu\left(\alpha_{i} - \sigma\left(\beta \right) \norm{\alpha}_{1} \right)}{\partial \alpha_i}.
\end{equation}
This approach leads to a learning mechanism similar to Discovering Neural Wiring~\cite{Wortsman2019}.
We will further discuss it in the section of the experiment and Appendix. 

\section{Application and Experiment}


We demonstrate the effectiveness of the proposed approach in learning a sparse structure.
It is important to note that our intend is not to compete with previous methods 
but rather to showcase the feasibility and potential of our proposed approach 
as the first fully differentiable sparsification method.

\subsection{Channel Pruning and Sparse Batch Normalization}

Network-slimming(NS)~\cite{Liu2019_pruning} prunes
unimportant channels in convolutional layers by
leveraging the scaling factors in batch normalization.
Let $x_i$ and $y_i$ be the input and output, respectively, of batch normalization for channel $i$.
Then, the operation can be written as
\begin{equation}\label{eqn:bn}
    \tilde{x}_{i} = \frac{x_i-\mu_{i}}{\sqrt{\sigma^{2}_{i}+\epsilon}};~y_i = a_i\tilde{x}_{i} + b_i,
\end{equation}
where $\mu_{i}$ and $\sigma_{i}$ denotes the mean and standard deviation of input activations, respectively,
$a_i$ and $b_i$ are scale and shift parameters, and $\epsilon$ is a small constant for numerical stability.
The affine transformation can be re-written as
\begin{equation}\label{eqn:sbn}
    y_i = a_i\left(\tilde{x}_i + b_i \right).
\end{equation}
By forcing $a_i$ to be zero, the corresponding channel can be removed.
NS trains a network with $l_1$-regularization on $a$
to identify insignificant channels.
After the initial training, channels with small values of $a$ are pruned and subsequently retrained.
The scale parameter $a_i$ can be regarded as an importance score or an architecture parameter.
In our approach, we parameterize the scale parameter using Eq.~(\ref{eqn:exc_param})
and train a network with $l_1$-norm on $a$ using SGD method without pruning and fine-tuning.

\subsubsection {Preliminary Experiment.}
We conduct preliminary experiments on CIFAR10/100~\cite{Alex2009} to validate our implementation. 
For our experiments, we utilize ResNet~\cite{He2016,He2016ID} as a base network architecture. 
We use a standard data augmentation scheme, which includes random shifting and flipping, following the approach used in~\cite{He2016,He2016ID}.
In NS, $\lambda$ in Eq.(\ref{eqn:loss}) is fixed to $10^{-5}$ and pruning ratios are varied. 
On the other hand, in our approach, we vary the value of $\lambda$ to induce different levels of sparsity. 
In both approaches, we applied $l^{2}_{2}$-regularization (weight decay) alongside $l_1$-regularization.
 
In NS, pre-trained models are obtained by training networks for $160$ epochs with an initial learning rate of $0.1$. The learning rate is divided by $10$ at $50\%$ and $75\%$ of the total number of training epochs. After the training process is complete, channels with small values of $a$ are removed, and a slimmed network is re-trained for an additional $160$ epochs with the same settings as in the initial training, but the learned weights are not re-initialized.
In our approach, networks are simply trained for $320$ epochs without fine-tuning or re-training. 
In NS, the scale factor is initialized as $0.5$. 
In our approach, we set $\alpha_i=0.5 \left(n+1\right)/n$ and $\beta = \sigma^{-1}\left( 1/\left(n^2+n\right)\right) = -\log \left(n^2 + n - 1\right)$ in Eq.(\ref{eqn:exc_param}) so that $a_i$ starts at $0.5$.

Table~\ref{tab:resnet_cifar10} presents the experimental results on CIFAR10. 
Experiments on CIFAR100 can be found in Appendix. 
Each experiment was run $5$ times, and the table displays the average and standard deviation from these repetitions.
Our proposed method is referred to as \emph{Differentiable Sparsification} (DS).
We controlled the value of $\lambda$ to achieve a similar computational cost as NS.
In the table, \emph{Channel} and \emph{Layer Sparsity} indicate the percentage of removed channels and layers, where larger values are preferable. 
It shows that the depth of DNNs can also be learned by removing redundant channels. 
The shortcuts across layers ensures that back-propagation is not impeded~\cite{Wen2016}.
NS has fewer chances of removing an entire layer. 
This can be attributed to the practice that NS simply prunes channels globally across all layers, 
without considering the inter-relationships among channels during the pruning procedure.


\subsubsection{Exploring Diverse Sparsity Patterns.}

\begin{table*}[!htbp]
\centering
\begin{small}
\begin{tabular}{*1c*2c*1c*2c*1c*2c*1c*2c*1c*2c}
\toprule
\multirow{2}{*}{Model}  &  \multicolumn{2}{c}{Channel Sparsity(\%)} & {} &  \multicolumn{2}{c}{Group Sparsity(\%)} & {} &  \multicolumn{2}{c}{Top-1 Error(\%)}  & {} & \multicolumn{2}{c}{FLOPs}  & {} & \multicolumn{2}{c}{Params} \\ \cmidrule{2-3}\cmidrule{5-6}\cmidrule{8-9}\cmidrule{11-12} \cmidrule{14-15}
{}                      &  Avg.            & Std.                   & {} &  Avg.         & Std.                    & {} &  Avg.                &  Std.          & {} & Avg.                 & Std. & {} & Avg.      & Std. \\
\midrule
Base        &  0.0            & 0.0            & {} &  0.0         & 0.0                 & {} & $5.92$   & $0.20$  & {} & $ 5.0\times10^{8} $   & 0.0 & {} & $ 1.1\times10^{6} $  & $ 0.0 $                                         \\
\midrule
DS-$l_1$        &  50.0          & 0.6            & {} & $1.7$         & 0.9             & {} & $5.72$   & $0.09$  & {} & $ 60.9\%$  & $ 0.9\% $ & {} & $55.2\%$   & $0.6\%$                               \\ 
DS-$l_{2,1}$    &  50.3            & 1.0            & {} & $45.8$         & 0.8             & {} & $5.79$   & $0.11$  & {} & $ 61.6\%$  & $ 0.9\% $ & {} & $55.5\%$   & $1.0\%$                               \\ 
\bottomrule
\end{tabular}
\end{small}
\caption{Performance comparison of $l_1$- and $l_{2,1}$-norm on CIFAR10, DenseNet-40-K12.}
\label{tab:densenet40}
\end{table*}

By simply switching from one norm to another for the regularization term, 
our method can achieve different sparsity patterns. 
To demonstrate this, we trained DenseNet-K$12$ with $40$ layers~\cite{Huang2017} 
with $l_1$-norm and $l_{2,1}$-norm on $a$ of Eq.(\ref{eqn:sbn}). 
In the $l_{2,1}$-norm case, we created a group of $12$ channels to learn layer-wise connections. 
In this experiment, we intentionally omitted weight decay on $\alpha_i$ and $\beta$ in Eq.(\ref{eqn:exc_param}) to isolate the effects of the $l_1$- and $l_{2,1}$-norms.

\begin{figure}[!t]
\centering
\begin{subfigure}{.23\textwidth}
  \centering
  \includegraphics[width=0.90\linewidth]{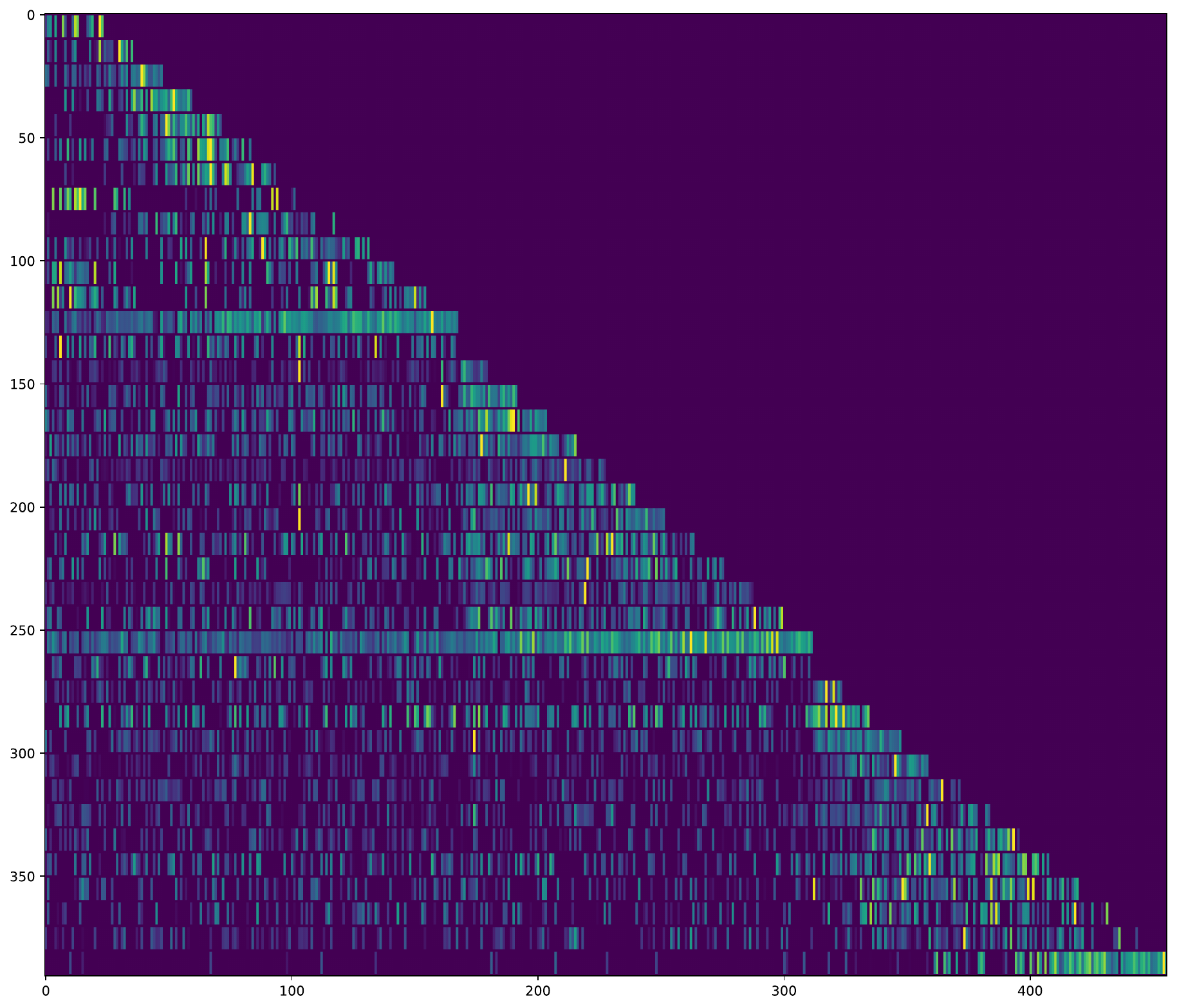}
  \caption{$l_1$-norm}
  \label{fig:sub-first}
\end{subfigure}
\begin{subfigure}{.23\textwidth}
  \centering
  \includegraphics[width=0.90\linewidth]{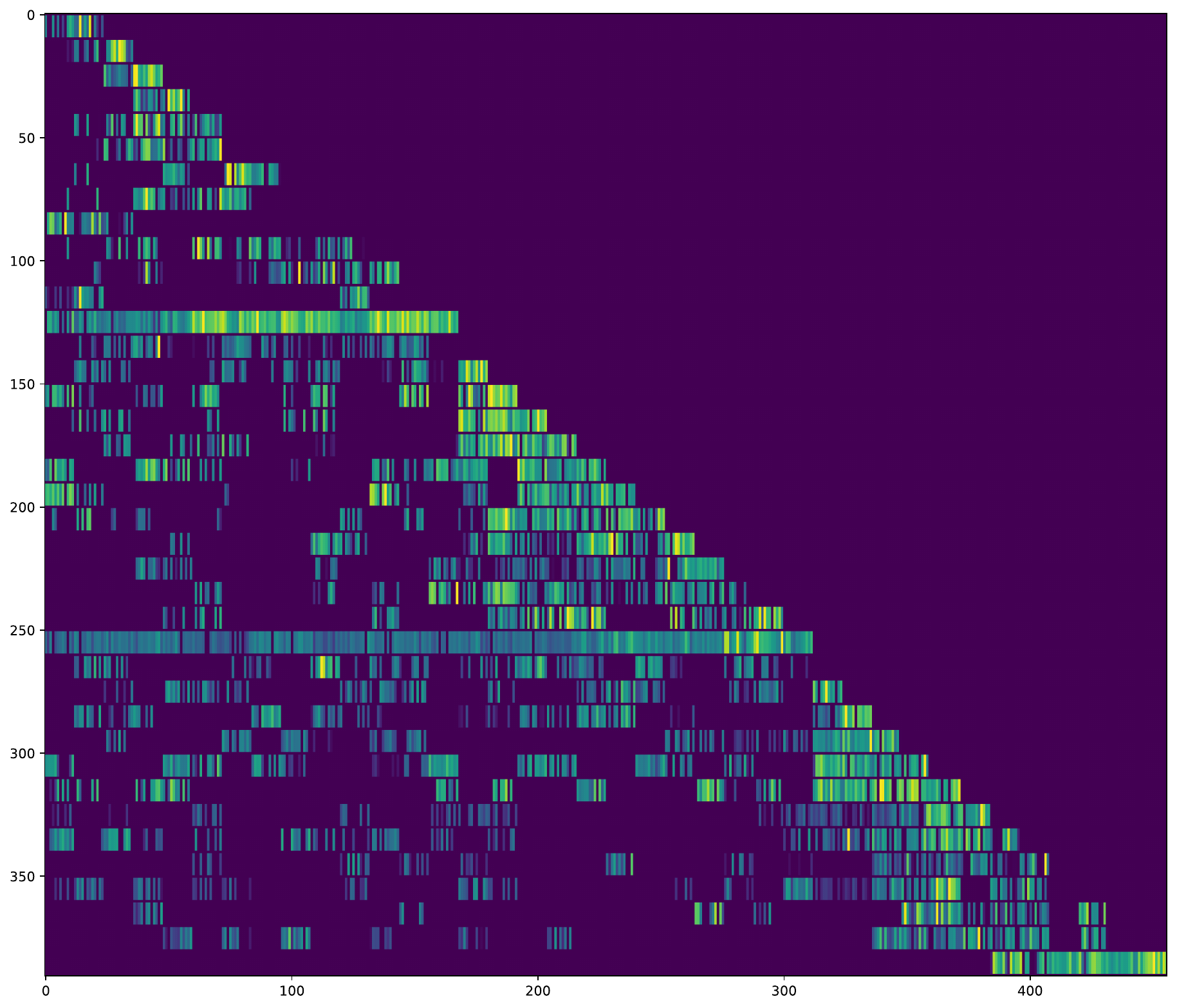}
  \caption{$l_{2,1}$-group norm}
  \label{fig:sub-second}
\end{subfigure}
\caption{Channel-wise view. DenseNet-40-K12.}
\label{fig:channel}
\end{figure}

\begin{figure}[!t]
\centering
\begin{subfigure}{.23\textwidth}
  \centering
  \includegraphics[width=0.90\linewidth]{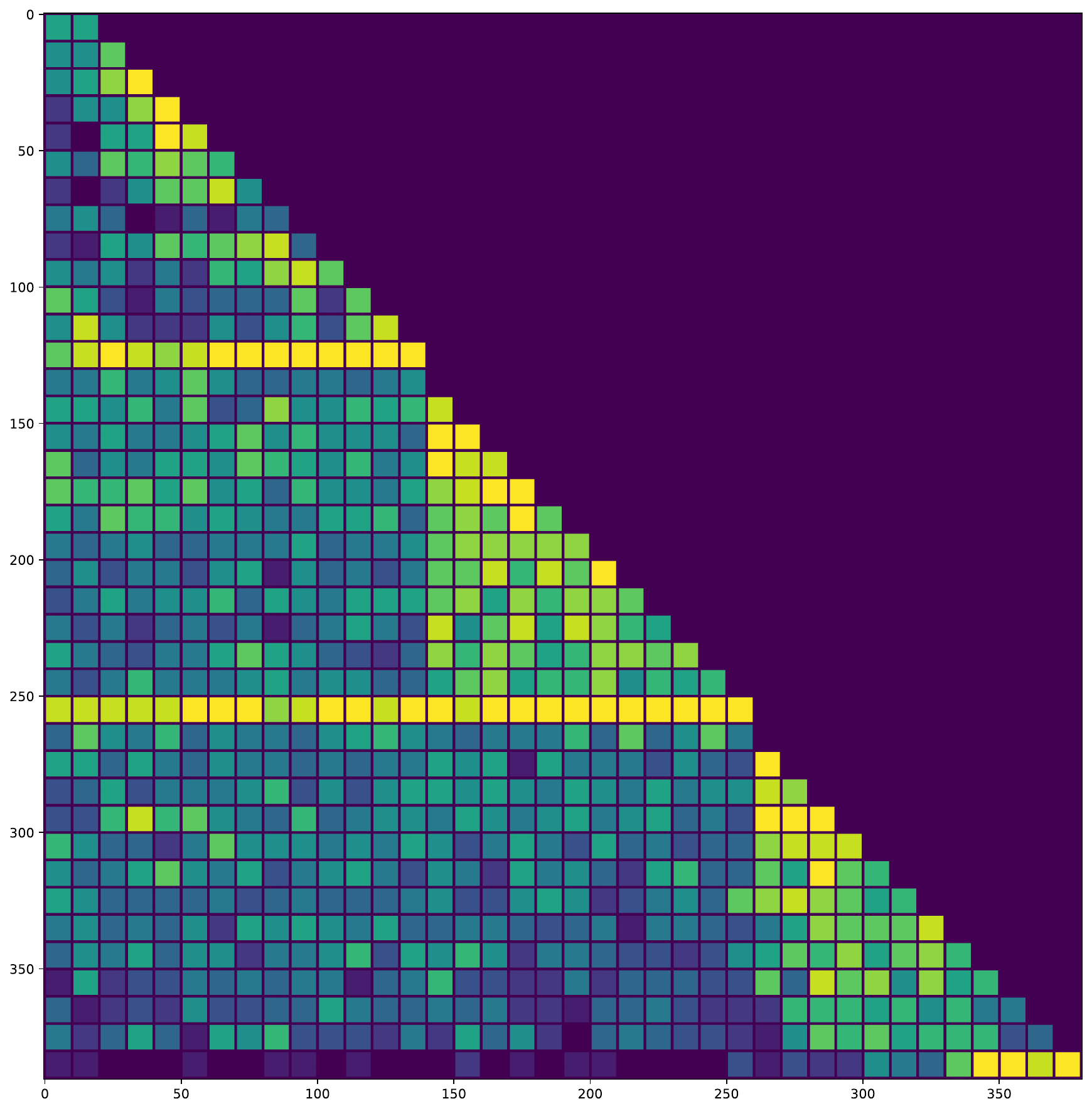}
  \caption{$l_1$-norm}
  \label{fig:sub-first}
\end{subfigure}
\begin{subfigure}{.23\textwidth}
  \centering
  \includegraphics[width=0.90\linewidth]{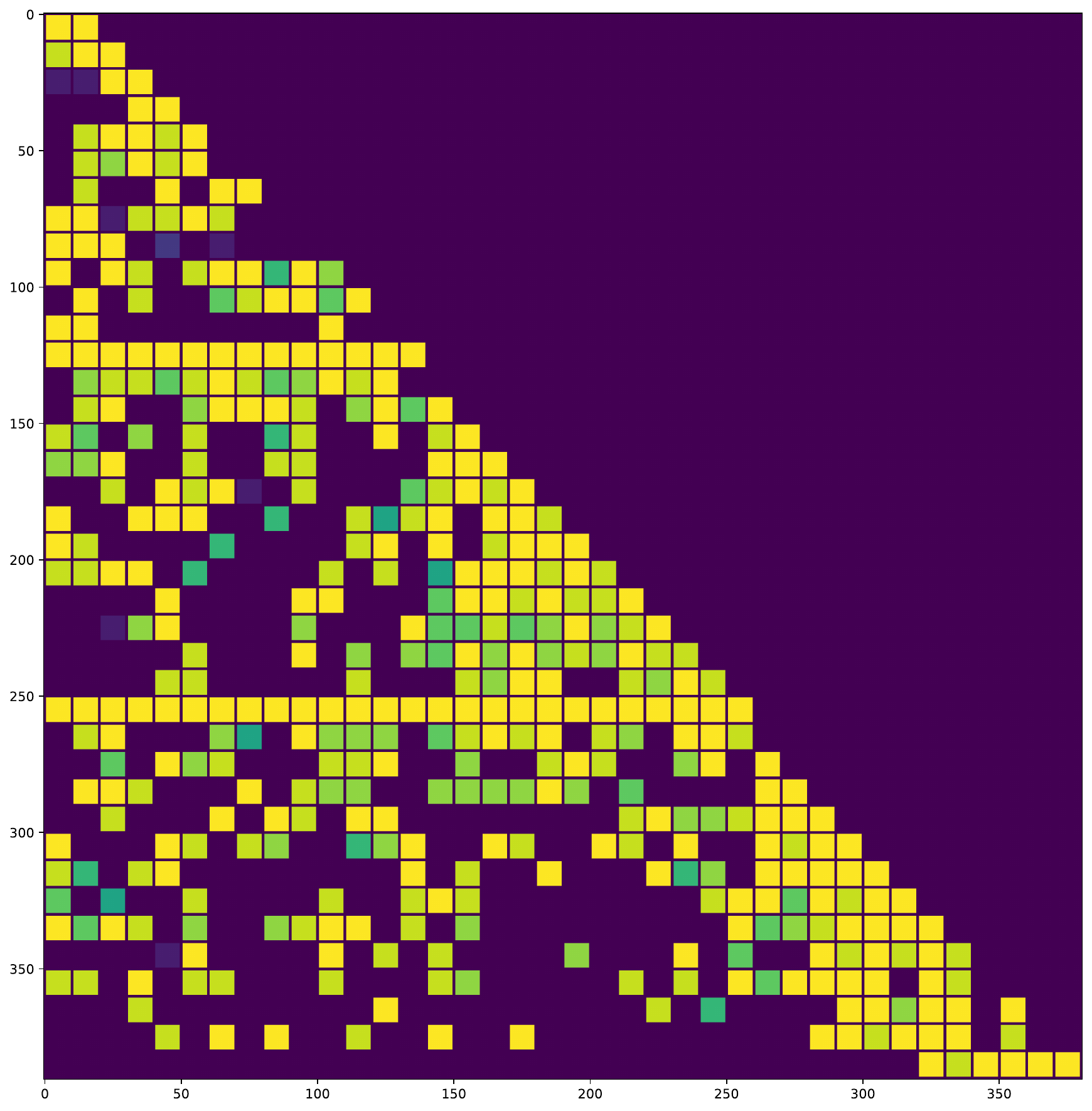}
  \caption{$l_{2,1}$-group norm}
  \label{fig:sub-second}
\end{subfigure}
\caption{Group-wise view. DenseNet-40-K12.}
\label{fig:group}
\end{figure}

Figures~\ref{fig:channel} and~\ref{fig:group} illustrate the learned sparse structures. Each row corresponds to a set of inputs to a hidden layer. Fig.\ref{fig:channel} shows the sparsified structures of the hidden layers in a channel-wise manner. 
The pixel-like thin short strip represents the magnitude of a scale parameter in the sparse batch normalization.
Fig.\ref{fig:group} shows the sparsified structures in a group-wise manner. Each square block in Fig.~\ref{fig:group} represents a group of $12$ channels. One block is added at a time as a layer proceeds from top to bottom since each hidden layer outputs $12$ new channels. An input layer generates $24$ channels, so the first row has $2$ blocks. The number of surviving channels is represented by the color of the block, and the brightness is proportionate to the number of non-zero channels within a group. 
The figures clearly illustrate that the $l_1$-norm acts on individual channels, 
and the removed channels are irregularly spread. 
On the contrary, the $l_{2,1}$-norm tends to either preserve an entire group or remove all the channels in a group.
Table~\ref{tab:densenet40} shows the sparsity rates and classification performances for the proposed method (DS).
The two experiments with $l_{1}$-norm and $l_{2,1}$-norm exhibit similar channel sparsity ratios, 
but their group sparsity ratios are notably different. 

\subsubsection{Knowledge Distillation.}
Knowledge distillation enables a student network to leverage a teacher's knowledge~\cite{Hinton2015distilling}. 
With the knowledge distillation, we can create a compact and high-performing model.
A distillation loss is used as the prediction loss in Eq.(\ref{eqn:loss}):
\begin{eqnarray}\label{eqn:kd_loss}
\mathcal{L} &=& \left(1-\alpha\right) \mathcal{L}_{ce}\left(\textbf{y}, \sigma\left(\textbf{z}_s \right)\right) \nonumber \\
& &+ \alpha T^2 \mathcal{L}{ce}\left(\sigma \left( \frac{\textbf{z}_s} {T} \right), \sigma \left( \frac{\textbf{z}_t} {T} \right)\right), 
\end{eqnarray}
where $\mathcal{L}{ce}$ denotes the cross-entropy loss, $\sigma\left( \cdot \right)$ is the softmax function, and $\textbf{z}_s$ and $\textbf{z}_t$ are the logit outputs of a student and a teacher network, respectively. A student network utilizes $l_1$-norm and the sparse batch normalization of Eq.(\ref{eqn:sbn}). 
On the other hand, a teacher network uses the normal batch normalization of Eq.(\ref{eqn:bn}). 

Following the practice of Hinge~\cite{Li2020}, 
we set a fixed balancing factor $\alpha=0.4$ and a temperature $T=4$ in Eq.(\ref{eqn:kd_loss}) and 
trained a teacher network on CIFAR$10/100$ and ImageNet~\cite{Deng2009ImageNet} for $320$ and $90$ epochs, respectively.
\citeauthor{Li2020} fine-tuned a student network for a maximum of $500$ epochs.
In our case, we trained a student network on CIFAR$10/100$ and ImageNet$2012$ for $400$ and $500$ epochs, respectively. 
For CIFAR$10/100$, we reduced the learning rate by a factor of $10$ at $240$ and $320$ epochs.
For ImageNet$2012$, we reduced the learning rate at $430$, $460$, and $490$ epochs and 
we gradually increased the value of $\lambda$ from an initial value of $\lambda_i=0$ to a final value of $\lambda_{f}$ over a span of $n=430$ epochs (see Appendix for details). 


\begin{table}
\centering
\begin{small}
\begin{tabular}{*1c*1c*1c*1c}
\toprule
   Method             &   Top-1 Error~/~BL(\%)  &FLOPs(\%)   &  Params(\%)  \\
\midrule

SSS\tablefootnote{\cite{Huang2018ECCV}, $^2$\cite{Li2020}, $^3$\cite{Zhao2019Var}, $^4$\cite{He2019FPGM}, $^5$\cite{Sui2021CHIP}, $^6$\cite{Hou2022CHEX} }  & 5.78~/~5.18   &  53.53 & 84.75 \\
Hinge\footnotemark[2]       & 5.40~/~4.97   &  53.61 & 70.34 \\
\midrule
\multirow{2}{*}{Ours}                    & 4.80~/~4.72   &  51.60 & 69.49 \\
{}                                       & 5.37~/~4.72   &  29.46 & 48.00 \\
\bottomrule
\end{tabular}
\end{small}
\caption{Knowledge Distillation. CIFAR10, ResNet-164.}
\label{tab:resnet164_kd_cifar10}
\end{table}

\begin{table}
\centering
\begin{small}
\begin{tabular}{*1c*1c*1c*1c}
\toprule
   Method             &   Top-1 Error~/~BL(\%)  &FLOPs(\%)   &  Params(\%)  \\
\midrule

Var\footnotemark[3]  & 6.84~/~5.89   &  55.22 & 40.33 \\
Hinge\footnotemark[2]       & 5.33~/~5.26   &  55.60 & 72.46 \\
\midrule
\multirow{2}{*}{Ours}                    & 4.96~/~5.32   &  52.27 & 51.32 \\
{}                                       & 5.32~/~5.32   &  36.81 & 33.27 \\
\bottomrule
\end{tabular}
\end{small}
\caption{Knowledge Distillation. CIFAR10, DenseNet-40.}
\label{tab:densenet40_kd_cifar10}
\end{table}

%

\begin{table}[!t]
\centering
\begin{small}
\begin{tabular}{*1c*1c*1c*1c}
\toprule
   Method             &   Top-1 Error~/~BL(\%)  &FLOPs(\%)  &  Params(\%)  \\
\midrule

SSS\footnotemark[1]                           & 25.82~/~23.88 &  68.92 & 72.94\\
Hinge\footnotemark[2]                         & 25.30~/~23.87 &  46.55 & -  \\ 
FPGM\footnotemark[4]                          & 25.17~/~23.85 &  46.50 & - \\ 
CHIP\footnotemark[5]                          & 23.85~/~23.85 &  51.30 & 55.80 \\ 
CHEX\footnotemark[6]                          & 22.60~/~22.20 &  48.78 & -  \\ 
\midrule
\multirow{2}{*}{Ours}                         & 24.73~/~24.47 &  60.06 & 65.91\\ 
{}                                            & 26.42~/~24.47 &  46.41 & 54.92\\ 
\bottomrule
\end{tabular}
\end{small}
\caption{Knowledge Distillation. ImageNet, ResNet-50.}
\label{tab:resnet50_kd_imagenet}
\end{table}

Tables~\ref{tab:resnet164_kd_cifar10},~\ref{tab:densenet40_kd_cifar10} and~\ref{tab:resnet50_kd_imagenet} present the experimental results 
on CIFAR$10$ and ImageNet$2012$.
Experimental results on CIFAR$100$ are given in Appendix. 
For CIFAR$10/100$, our proposed method was tested five times, and the reported values in the tables are averages of these five repetitions. 
For ImageNet$2012$, each experiment was conducted once. 
These tables provide the remaining percentage of FLOPs and parameters for the sparsified or pruned models, where lower values are preferable. 
The performance metrics of other methods are referenced from their original works. 
It is important to note that due to varying compression ratios and different learning settings across methods, direct comparisons might not be straightforward. 
Hence, the presented results should be interpreted as reference points. 
Despite its relatively simple concept and implementation, our approach demonstrates comparable performances with other methods.

\subsection{Discovering Neural Wiring}

\begin{table*}[!htbp]
\centering
\begin{small}
\begin{tabular}{*1c*2c*1c*2c*1c*2c}
\toprule
\multirow{2}{*}{Model}  & \multicolumn{2}{c}{Top-1 Error(\%)} & {} & \multicolumn{2}{c}{Mult-Adds}  & {} &  \multicolumn{2}{c}{Parmas}              \\\cmidrule{2-3}\cmidrule{5-6}\cmidrule{8-9}
{}                      &  Avg.           & Std.              & {} &  Avg.                &  Std.             & {} & Avg.                 & Std. \\
\midrule
\multicolumn{1}{l}{MobileNetV1($\times0.25$)}& 13.44          & 0.24              & {} & $ 3.3\times10^{6} $   & 0.0 & {} & $2.2\times10^{5}$   & 0.0               \\
\midrule
\multicolumn{1}{l}{DNW-No Update($\times0.225$)}&13.86         & 0.27             & {} & $ 4.5\times10^{6} $   & $ 3.7\times10^{4} $ & {} & $2.2\times10^{5}$   & $3.7\times10^{1}$               \\
\multicolumn{1}{l}{DNW-Update($\times0.225$)}    &10.30         & 0.20            & {} & $ 3.1\times10^{6} $   & $ 4.6\times10^{4} $ & {} & $1.8\times10^{5}$   & $6.7\times10^{1}$               \\
\midrule
\multicolumn{1}{l}{PG-$l_{1}$-norm}  & 12.17       & 0.44                         & {} & $ 3.3\times10^{6} $   & $ 1.7\times10^{5} $ & {} & $2.1\times10^{4}$   & $9.4\times10^{2}$ \\
\multicolumn{1}{l}{PG-$l_{1,2}$-norm} & 13.62       & 0.56                        & {} & $ 3.4\times10^{6} $   & $ 8.6\times10^{4} $ & {} & $9.6\times10^{4}$   & $1.6\times10^{4}$ \\
\midrule
\multicolumn{1}{l}{DS-No RGF}    & 10.55         & 0.23                           & {} & $ 3.4\times10^{6} $   & $ 4.5\times10^{4} $ & {} & $6.1\times10^{4}$   & $5.7\times10^{2}$     \\
\multicolumn{1}{l}{DS-RGF}      & 9.36         & 0.27                             & {} & $ 3.3\times10^{6} $   &$ 6.7\times10^{4} $ & {} & $4.7\times10^{4}$   & $8.4\times10^{2}$        \\
\bottomrule
\end{tabular}
\end{small}
\caption{Discovering Neural Wiring. Performance on CIFAR-10}
\label{tab:dnw_cifar10}
\end{table*}

Discovering Neural Wiring (DNW)~\cite{Wortsman2019}
relaxes the notion of layers and treats channels as nodes in a neural graph.
By learning connections between nodes, DNW simultaneously discovers the structure and learns the parameters of a neural graph.
The input to node $v$, denoted as $\textbf{y}_{v}$, is expressed as:
\[
    \textbf{y}_{v} = \sum_{\left( u,v\right) \in \mathcal{E}} w_{u,v}\textbf{x}_{u},
\] where $\textbf{x}_{u}$ denotes the state of a preceding node,
$\mathcal{E}$ represents an edge set and $w_{u,v}$ is the connection weight of an edge.
The neural graph's structure is determined by selecting a subset of edges. 
During each iteration of the training process, DNW identifies the top $k$ edges with the highest magnitudes, named as the real edge set, 
while the remaining edges constitute the hallucinated edge set. 
During the forward pass, only real edges are utilized. 
DNW allows the weights in both sets to change throughout training, enabling a hallucinated edge to replace a real edge if it becomes strong enough.
While the weights of real edges are updated using standard SGD, those of hallucinated edges follow a specialized learning rule: the gradient flows to hallucinated edges, but it does not pass through them. 
This distinction ensures that the hallucinated edges contribute to the network's architecture discovery without directly affecting the other parameter updates.

In our proposed method, the architecture parameters correspond to the edges in DNW. We parameterize these edges using Eq.~(\ref{eqn:exc_param}),
\begin{equation}\label{eqn:dnw_exc_param}
w_{u,v} = sign\left(w_{u,v}\right) \left(\abs{w_{u,v}} - \sigma\left(\beta \right) \norm{w_{:,v}}_{1}\right)_{+}, \nonumber
\end{equation}
 and train the network with $l_1$-norm on the edges to induce sparsity.
Along with the sparse parametrization, the rectified gradient flow (RGF) of Eq.(\ref{eqn:rgf}) yields an update rule which is similar to that of DNW. 
However, unlike DNW, there is no need to keep track of real and hallucinated edge sets. 
Instead, we directly optimize an objective function using the approximated gradients.
Implementing RGF is straightforward and the reference code is provided in the supplementary material.

We conduct experiments on CIFAR10/100 and ImageNet2012. 
Our base model is MobileNetV$1$~\cite{Andrew2017}, and our implementation closely follows that of DNW. 
The model is trained for $160$ and $250$ epochs on CIFAR10/100 and ImageNet2012, respectively. 
An initial learning rate is set to $0.1$ and scheduled using cosine annealing. 
To determine the sparsity level for DNW and our proposed method, we vary the value of $k$ and $\lambda$, respectively. 
For RGF, the hyper-parameter of \emph{elu}, which controls the extents of saturation for negative inputs, is set to $0.1$.

Table~\ref{tab:dnw_cifar10} presents the experimental results on CIFAR10.  
Experimental results on CIFAR100 are given in Appendix. 
Our approach is denoted as DS. We also include results for DS without RGF, which corresponds to DNW-No Update, where the magnitudes of hallucinated edges are not updated. 
Even without RGF, the performance of our proposed method is close to that of DNW with the update rule. 
We further conducted experiments using proximal gradients in Eq.(\ref{eqn:l1_prox}) and Eq.(\ref{eqn:exc_prox}), denoted as PG in Table~\ref{tab:dnw_cifar10}. PG-$l_1$-norm also employs the $l_{1}$-norm as a regularizer, but it did not achieve the same level of performance as our approach. PG-$l_{1,2}$-norm uses the update rule of Eq.(\ref{eqn:exc_prox}), which has a similar shape to our proposed sparse parameterization of Eq.~(\ref{eqn:exc_param}), but its performance is inferior to ours. Table~\ref{tab:dnw imagenet} presents the experimental results on ImageNet2012. 

\begin{table}[]
\centering
\begin{small}
\begin{tabular}{*1c*1c*1c*1c}
\toprule
   Method                                        &   Top-1 Error  &  Mult-Adds &  Params  \\
\midrule
MobileNetV1\tablefootnote[7]{\cite{Andrew2017},~$^{8}$\cite{Ma2018ShuffleNetV2},~$^{9}$\cite{Wortsman2019}}  & 36.30           &  $149M$    & $1.30M$   \\ 
ShuffleNetV2\footnotemark[8]                   & 30.60           &  $146M$    & $2.30M$  \\ 
MobiletNetV1-DNW\footnotemark[9] & 29.60           &  $154M$    & $1.80M$ \\ 
\midrule
\multirow{2}{*}{Ours}                         & 29.45          &  $154M$    & $1.62M$ \\
{}                                            & 30.20          &  $147M$    & $1.56M$ \\ 
\bottomrule
\end{tabular}
\end{small}
\caption{Discovering Neural Wiring. ImageNet.}
\label{tab:dnw imagenet}
\end{table}


\subsection{Learning Sparse Affinity Matrix}

\begin{figure}[]
\centering
\includegraphics[width=0.925\columnwidth]{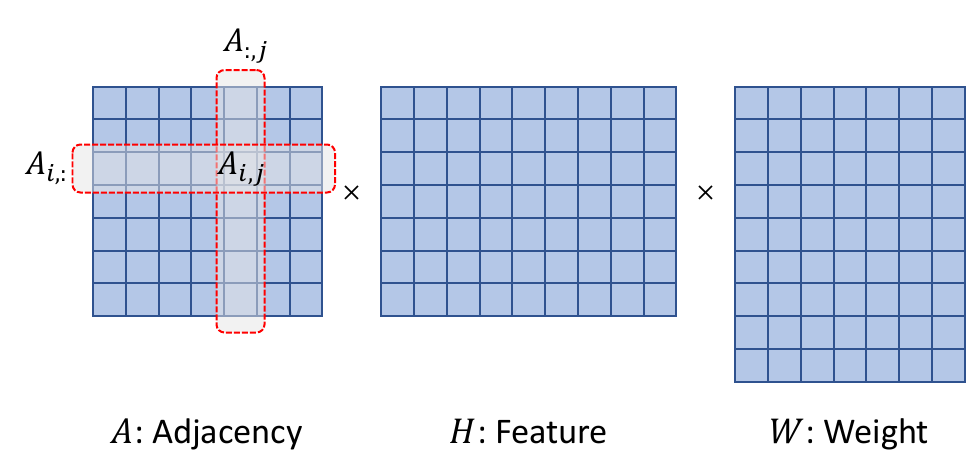}
\caption{Structure of a GCN block. Each block consists of a shared adjacency, an input feature, and a weight matrix.
Each row and column of an adjacent matrix are treated as groups to learn the relationships between a node and its neighbors.}
\label{fig:gcn}
\end{figure}

We employ the proposed approach to learn the sparse structure of an adjacency matrix in a graph convolutional network (GCN)~\cite{Kipf2017}.
A GCN block or layer is defined (see Fig.~\ref{fig:gcn}) as
\[ H^{l+1} = F \left( A H^{l} W^{l} \right), \]
where $A$ is an adjacency matrix; $H^{l}$ and $W^{l}$ are an input feature and a weight matrix for layer $l$, respectively;
and $F$ is a nonlinear activation function.
In general, $A$ is non-negative and shared across GCN blocks.
It is obtained by normalization.
For example, $A = \tilde{D}^{-1} \tilde{A} $ or $A = \tilde{D}^{-\frac{1}{2}} \tilde{A} \tilde{D}^{-\frac{1}{2}}$,
where $\tilde{A}$ is an unnormalized adjacency matrix;
and $\tilde{D}$ is a diagonal matrix, where $\tilde{D}_{i}=\sum_{j} \tilde{A}_{i,j}$.
The adjacency matrix represents the connections or relationships between nodes on a graph and
is usually given by prior knowledge.
Learning the value of $A_{i,j}$ amounts to determining the relationship between nodes $i$ and $j$.
If the value of $A_{i,j}$ is zero, the two nodes are thought to be unrelated.

As shown in Fig.~\ref{fig:gcn},
we define each row and column as a group.
Row grouping creates competition between in-coming edges,
whereas column grouping creates competition between out-going edges.
Each row and column of unnormalized adjacency matrix $\tilde{A}$ can be parameterized similarly
as in $\tilde{\gamma}$ of in Eq.~(\ref{eqn:thresholding}):
\[
\tilde{A}_{i,j} = \left( \gamma_{i,j} - \sigma\left( \beta^r_i \right) \cdot \norm{\gamma_{i,:}}_{1} - \sigma\left( \beta^c_j \right) \cdot \norm{\gamma_{:,j}}_{1} \right)_{+},
\] where $ \gamma_{i,j} = \exp\left(\alpha_{i,j}\right).$
To convert $\tilde{A}$ into a doubly stochastic matrix, we employ an iterative balanced normalization, which can be considered as a variant of Sinkhorn normalization \cite{Sinkhorn1964,Knopp1967,Knight2008}. 
We iteratively apply the following equations after initializing $A$ with $\tilde{A}$:
\[
A = D_{r}^{-\frac{1}{2}} A D_{c}^{-\frac{1}{2}}, 
\] where $D_{r}$ and $D_{c}$ are diagonal matrices with $\left[D_r\right]_i = \sum_j A_{i,j}$ and $\left[D_c\right]_j = \sum_i A_{i,j}$, respectively.
Despite being an iterative process, this normalization is differentiable. 
Through numerical experiments, we have observed that the balanced normalization indeed makes $\tilde{A}$ doubly stochastic. However, we currently lack a theoretical justification for this observation, leaving the mathematical proof as an open question for future research.

As competition groups are created in row- and column-wise approaches, a regularized objective function can be written as
\[
  \mathcal{L} \left(  D, W, A \right) + \frac{\lambda}{2} \sum_{i=1}^{N} \left\{ \mathcal{R} \left(A_{i,:} \right) + \mathcal{R} \left(A_{:,i} \right) \right\},
\]
where $W=\{W^{l}\}$, $N$ is the size of square matrix $A$, and $A_{i,:}$ and $A_{:,i}$ denote $i$th row and column vector of $A$, respectively.
We employ the $l_p$-norm (Eq.~(\ref{eqn:lp_norm})) with $p=0.5$ as the regularization function $\mathcal{R}$, which encourages the sparsity and competition within the groups.
Note that $l_1$-norm may not work for this since the sum of a column and a row is equal to one.


To validate our proposed method, we applied the proposed GCN for traffic speed prediction in a road network. The traffic speed data were collected from $170$ road segments, resulting in an adjacency matrix of size $170 \times 170$. 
We predict a one-step-ahead observation from the previous eight observations. 
The output layer generates $170$ estimates, one for each road segment. We employed the mean relative error (MRE) as the prediction loss for training. 
Additional details regarding the experimental data and our GCN model can be found in Appendix.


\begin{table}[!tbp]
\centering
\begin{small}
\begin{tabular}{*1c*1c*1c*2r}
\toprule
\multirow{2}{*}{Model} & \multirow{2}{*}{\# Non-zeros}& \multirow{2}{*}{MAPE(\%)} & \multicolumn{2}{c}{LR \scriptsize{($\times100$)}} \\\cmidrule{4-5}
    {}                 &   {}                         &    {}                     &  $k=1$         &        $k=2$             \\
\midrule
B.S. I        & 878   &  5.5160 & 100.00  & 100.00  \\
B.S. II       & 28,900 &  5.6343 &  13.76  &  20.87  \\
\midrule
\multirow{2}{*}{Ours} &  1,009   & 5.4957  & 88.62  & 91.39 \\
{}       &    835   & 5.5336  & 89.79  & 92.13  \\
\bottomrule
\end{tabular}
\end{small}
\caption{Traffic speed prediction with GCN.}
\label{tab:gcn}
\end{table}


Two baseline models were used. 
For the first baseline, we set $\tilde{A}_{i,j} = \exp\left(\alpha_{i,j}\right)$ if node $i$ and $j$ were adjacent to each other, 
$\tilde{A}_{i,j}=0$ otherwise. 
For the second baseline, no connectivity information was given. 
We set $\tilde{A}_{i,j} = \exp\left(\alpha_{i,j}\right)$ for all $i,j$, regardless of the actual connections.
For the proposed method, we parameterized the adjacency matrix with the sparsification technique and employed $l_{0.5}$-norm as a regularizer. 
Balanced normalization was applied to all cases.
To measure the learned relationship (LR) between nodes,
we propose the following scoring function:
\[
\frac{1}{2N} \sum^{N}_{i=1} \sum^{N}_{j=1} \left[ \left(A^r +A^c\right) \odot M^k\right]_{i,j},
\]
where $A^{r} = D^{-1}_{r}A$, $A^{c} = AD^{-1}_{c}$, $\odot$ denotes the element-wise product,
and $[ M^k ]_{i,j} = 1$ if the geodesic distance between node $i$ and $j$ is less than or equal to $k$,
whereas $[ M^k ]_{i,j} = 0$ otherwise.
The maximum value is $1$, and the minimum is $0$.
For example, the first baseline always takes the maximum value
because its adjacency matrix has exactly the same structure as $M^1$.
We calculated the scores for $k = 1$ and $2$
.

The experimental results are presented in Table~\ref{tab:gcn}. 
Each experiment was conducted five times, and the median among the five lowest validation errors was selected for analysis.
In Baseline I, road connectivity information is provided, and the number of non-zero elements in the adjacency matrices is $878$. 
For Baseline II, road connectivity information is not given, resulting in $28,900 (= 170 \times 170)$ non-zero elements. 
Baseline II learned a mapping between input and target values by solely reducing the prediction loss and failed to learn the relationships between nodes. 
In contrast, the proposed model learned sparse relationships between nodes.

\section{Conclusion}
In this study, we proposed a fully differentiable sparsification method that can simultaneously learn the sparsified structure and weights of deep neural networks.
Our proposed method is versatile in that it can seamlessly be integrated into different types of neural networks and
used to address a wide range of problems and scenarios.

\bibliography{reference,aaai24}

\end{document}



\onecolumn
 
\section{Appendix}

\subsection{Gradual Sparsity Regularization}
In the early stages of training,
it is difficult to determine the importance of parameters because they are randomly initialized.
A gradual scheduling of $\lambda$ provides a warm-up stage that prevents premature parameter dropping.
Inspired by the gradual pruning of~\cite{Zhu2017},
we smoothly increase the value of $\lambda$ from an initial value $\lambda_i$
to a final value $\lambda_f$ over a span of $n$ epochs.
Starting at epoch $t_0$, we increase the value at every epoch using the following formula:
\begin{equation}\label{eqn:gsr}
\lambda_t = \lambda_f + \left( \lambda_i - \lambda_f \right) \left( 1 - \frac{t-t_0}{n} \right)^3,
\end{equation} where $t$ denotes an epoch index.

\subsection{Channel Pruning and Sparse Batch Normalization}

\subsubsection{Preliminary Experiment.}

Tables~\ref{tab:resnet_cifar10} and \ref{tab:resnet_cifar100} present additional experimental results on CIFAR$10/100$~\cite{Alex2009}. 
In Network-slimming (NS)~\cite{Liu2019_pruning}, $\lambda$  is fixed to $10^{-5}$ and pruning ratios are varied. 
Conversely, in our approach (DS), we vary the value of $\lambda$ to achieve different degrees of sparsity. 
In the tables, we ranged the values of $\lambda$ from $0.0000150$ to $0.0000475$ for CIFAR10 and from $0.0000425$ to $0.0001250$ for CIFAR100.
In both approaches, we utilized $l^{2}_{2}$-regularization (weight decay) in conjunction with $l_1$-regularization on the scaling factors or architecture parameters. 
The weight of $l^{2}_{2}$-regularization is set to $0.00005$ for the scaling factor $a$ in NS and $0.00001$ for 
the free architecture parameters $\alpha_i$ and $\beta$ in our method (DS).

\begin{table*}[!h]
\centering
\begin{small}
\begin{tabular}{*1c*2c*1c*2c*1c*2c*1c*2c*1c*2c}
\toprule
\multirow{2}{*}{Model}  &  \multicolumn{2}{c}{Channel Sparsity(\%)} & {} &  \multicolumn{2}{c}{Layer Sparsity(\%)} & {} &  \multicolumn{2}{c}{Top-1 Error(\%)} & {} & \multicolumn{2}{c}{FLOPs}        & {} &  \multicolumn{2}{c}{Parmas}   \\\cmidrule{2-3}\cmidrule{5-6}\cmidrule{8-9}\cmidrule{11-12} \cmidrule{14-15}
{}                      &  Avg.            & Std.                   & {} &  Avg.         & Std.                    & {} &  Avg.         & Std.                 & {} &  Avg.                &  Std.     & {} & Avg.                 & Std. \\
\midrule
Base                    &  0.0             & 0.0                    & {} &  0.0          & 0.0                     & {} &  5.13         & 0.05                 & {} & $ 5.0\times10^{8} $  & 0.0       & {} & $1.7\times10^{6}$  & 0.0      \\
\midrule
\multirow{5}{*}{NS}     &  55.0            & 0.0                    & {} &  0.1          & 0.3                     & {} &  5.10         & 0.15                 & {} & $ 62.5\%$            & $ 0.7\% $ & {} &  $73.3\%$           & $0.6\%$  \\
{}                      &  60.0            & 0.0                    & {} &  0.1          & 0.3                     & {} &  5.10         & 0.16                 & {} & $ 56.3\%$            & $ 0.2\% $ & {} &  $66.6\%$           & $0.3\%$ \\
{}                      &  65.0            & 0.0                    & {} &  0.0          & 0.0                     & {} &  5.38         & 0.21                 & {} & $ 49.3\%$            & $ 0.4\% $ & {} &  $59.0\%$           & $0.4\%$  \\
{}                      &  70.0            & 0.0                    & {} &  0.0          & 0.0                     & {} &  5.59         & 0.09                 & {} & $ 41.6\%$            & $ 0.5\% $ & {} &  $50.0\%$           & $0.3\%$  \\
{}                      &  75.0            & 0.0                    & {} &  2.2          & 1.1                     & {} &  6.54         & 0.18                 & {} & $ 33.4\%$            & $ 0.6\% $ & {} &  $40.3\%$           & $0.5\%$  \\
\midrule               
\multirow{5}{*}{DS}     &  56.6            & 0.3                    & {} &  2.6          & 1.5                     & {} &  5.08         & 0.06                 & {} & $ 56.5\%$            & $0.8\%$   & {} &  $68.0\%$           & $0.8\%$  \\ 
{}                      &  60.5            & 0.3                    & {} &  5.0          & 1.4                     & {} &  5.32         & 0.16                 & {} & $ 51.6\%$            & $0.8\%$   & {} &  $63.2\%$           & $1.0\%$  \\
{}                      &  65.1            & 0.4                    & {} & 11.7          & 1.0                     & {} &  5.43         & 0.09                 & {} & $ 44.3\%$            & $0.8\%$   & {} &  $57.2\%$           & $0.4\%$  \\
{}                      &  70.3            & 0.4                    & {} & 21.2          & 4.2                     & {} &  5.48         & 0.17                 & {} & $ 37.4\%$            & $0.9\%$   & {} &  $50.1\%$           & $0.2\%$  \\
{}                      &  74.1            & 1.1                    & {} &  32.6         & 5.4                     & {} &  5.69         & 0.10                 & {} & $ 31.8\%$            & $0.9\%$   & {} &  $44.7\%$           & $0.7\%$  \\
\bottomrule
\end{tabular}
\end{small}
\caption{Performance on CIFAR-10, ResNet-164.}
\label{tab:resnet_cifar10}
\end{table*}

\begin{table*}[!h]
\centering
\begin{small}
\begin{tabular}{*1c*2c*1c*2c*1c*2c*1c*2c*1c*2c}
\toprule
\multirow{2}{*}{Model}  &  \multicolumn{2}{c}{Channel Sparsity(\%)} & {} &  \multicolumn{2}{c}{Layer Sparsity(\%)} & {} &  \multicolumn{2}{c}{Top-1 Error(\%)} & {} & \multicolumn{2}{c}{FLOPs}          & {} &  \multicolumn{2}{c}{Parmas}   \\\cmidrule{2-3}\cmidrule{5-6}\cmidrule{8-9}\cmidrule{11-12} \cmidrule{14-15}
{}                      &  Avg.            & Std.                   & {} &  Avg.         & Std.                    & {} &  Avg.         & Std.                 & {} &  Avg.               &  Std.   & {} & Avg.                 & Std. \\
\midrule
Base                    &  0.0             & 0.0                    & {} &  0.0          & 0.0                     & {} & 23.22         & 0.52                 & {} & $ 5.0\times10^{8} $  & 0.0    & {} &  $1.7\times10^{6}$  & 0.0      \\
\midrule
\multirow{5}{*}{NS}     &  50.0            & 0.0                    & {} &  0.3          & 0.3                     & {} & 22.78         & 0.31                 & {} & $ 62.5\%$            & $ 0.4\% $   & {} &  $78.7\%$           & $0.4\%$ \\
{}                      &  55.0            & 0.0                    & {} &  0.9          & 0.6                     & {} & 22.91         & 0.20                 & {} & $ 57.0\%$            & $ 1.0\% $   & {} &  $74.2\%$           & $0.4\%$  \\
{}                      &  60.0            & 0.0                    & {} &  0.0          & 0.0                     & {} & 23.86         & 0.28                 & {} & $ 50.9\%$            & $ 0.7\% $   & {} &  $69.5\%$           & $0.2\%$ \\
{}                      &  65.0            & 0.0                    & {} &  2.5          & 1.6                     & {} & 25.07         & 0.40                 & {} & $ 44.6\%$            & $ 0.7\% $   & {} &  $64.4\%$           & $0.4\%$  \\
{}                      &  70.0            & 0.0                    & {} &  3.4          & 1.1                     & {} & 26.80         & 0.23                 & {} & $ 37.8\%$            & $ 0.8\% $   & {} &  $58.0\%$           & $0.3\%$ \\
\midrule
\multirow{5}{*}{DS}     &  51.3            & 0.3                    & {} &  10.2         & 2.1                     & {} & 23.77         & 0.24                 & {} & $ 54.5\%$            & $1.5\%$     & {} &  $77.7\%$           & $0.2\%$  \\
{}                      &  56.1            & 0.4                    & {} &  14.5         & 1.3                     & {} & 23.83         & 0.25                 & {} & $ 49.7\%$            & $0.8\%$     & {} &  $74.1\%$           & $0.6\%$ \\
{}                      &  60.4            & 0.4                    & {} &  22.1         & 2.4                     & {} & 24.23         & 0.19                 & {} & $ 44.6\%$            & $1.4\%$     & {} &  $70.1\%$           & $0.6\%$  \\
{}                      &  66.1            & 0.4                    & {} &  34.6         & 2.5                     & {} & 24.52         & 0.22                 & {} & $ 37.3\%$            & $1.1\%$     & {} &  $63.6\%$           & $0.4\%$ \\
{}                      &  70.1            & 0.4                    & {} &  39.1         & 1.8                     & {} & 25.28         & 0.39                 & {} & $ 32.9\%$            & $0.7\%$     & {} &  $56.9\%$           & $0.4\%$ \\

\bottomrule
\end{tabular}
\end{small}
\caption{Performance on CIFAR-100, ResNet-164.}
\label{tab:resnet_cifar100}
\end{table*}

\subsubsection{Exploring Diverse Sparsity Patterns.}

\begin{figure}[!th]
\centering
\begin{subfigure}{.30\textwidth}
  \centering
  \includegraphics[width=0.90\linewidth]{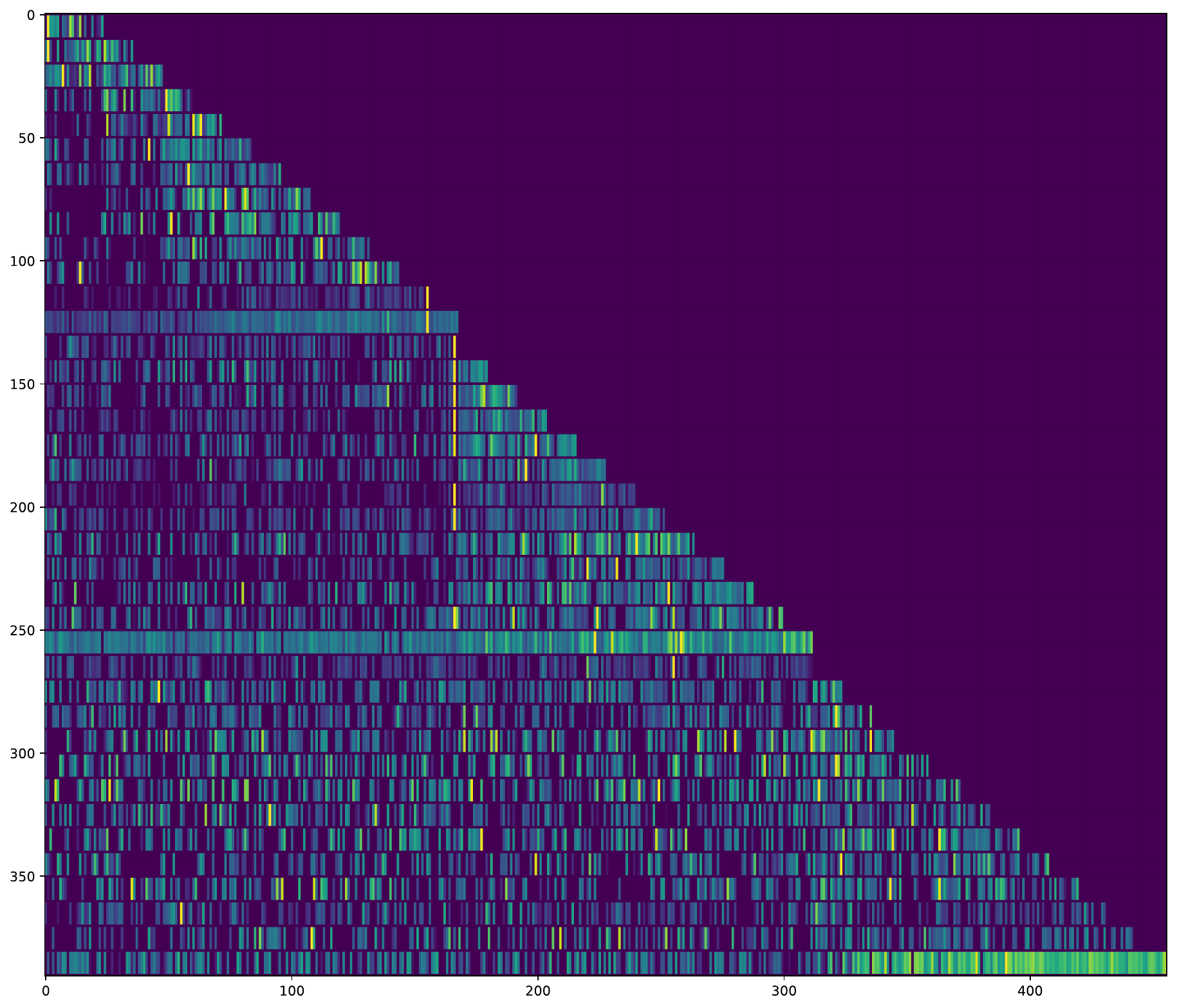}
  \caption{$l_1$-norm}
  \label{fig:sub-first}
\end{subfigure}
\begin{subfigure}{.30\textwidth}
  \centering
  \includegraphics[width=0.90\linewidth]{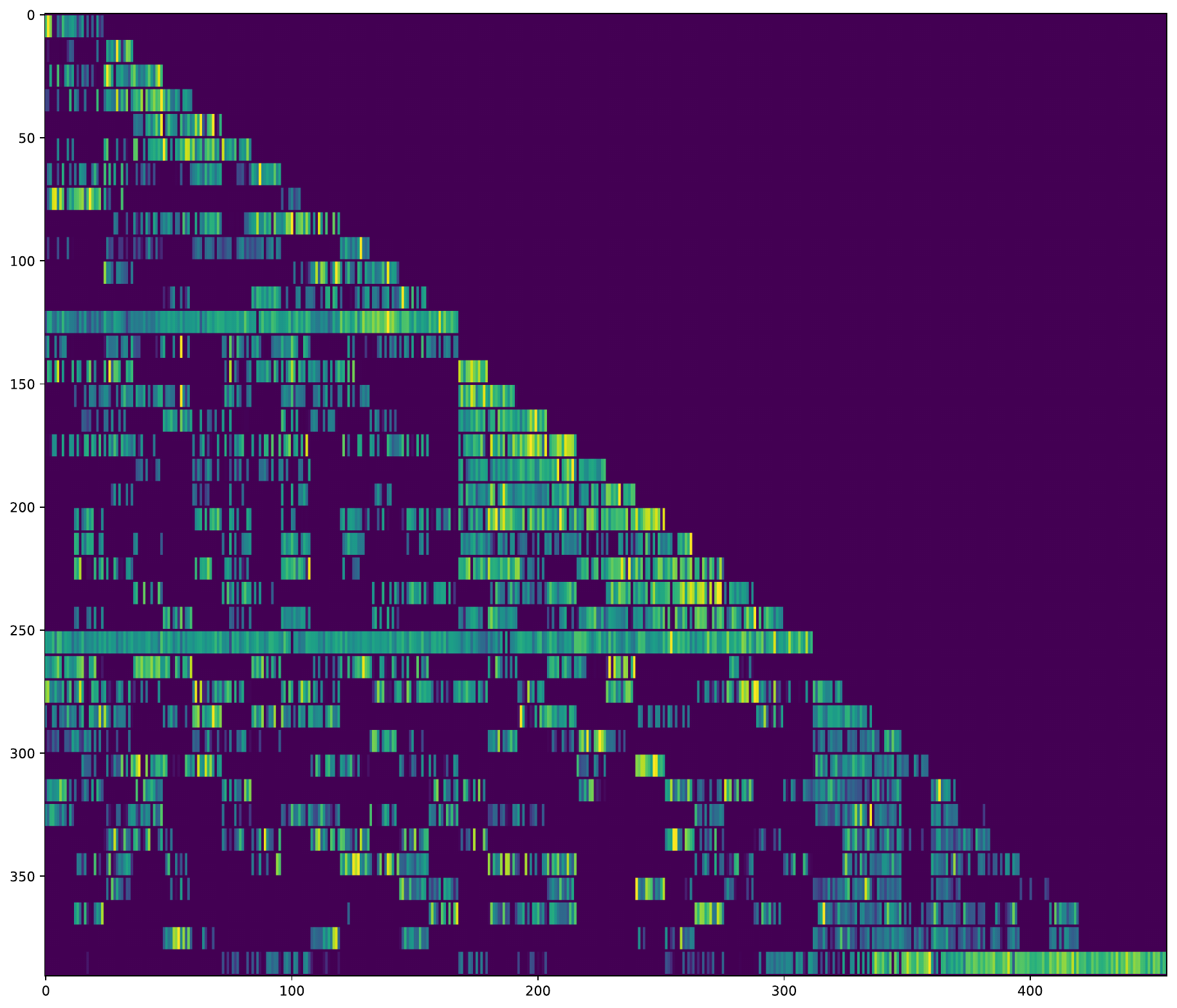}
  \caption{$l_{2,1}$-group norm}
  \label{fig:sub-second}
\end{subfigure}
\caption{Sparsified Connections in Channel-wise view. DenseNet-40-K12. CIFAR100.}
\label{fig:cifar100_channel}
\end{figure}

\begin{figure}[!th]
\centering
\begin{subfigure}{.30\textwidth}
  \centering
  \includegraphics[width=0.90\linewidth]{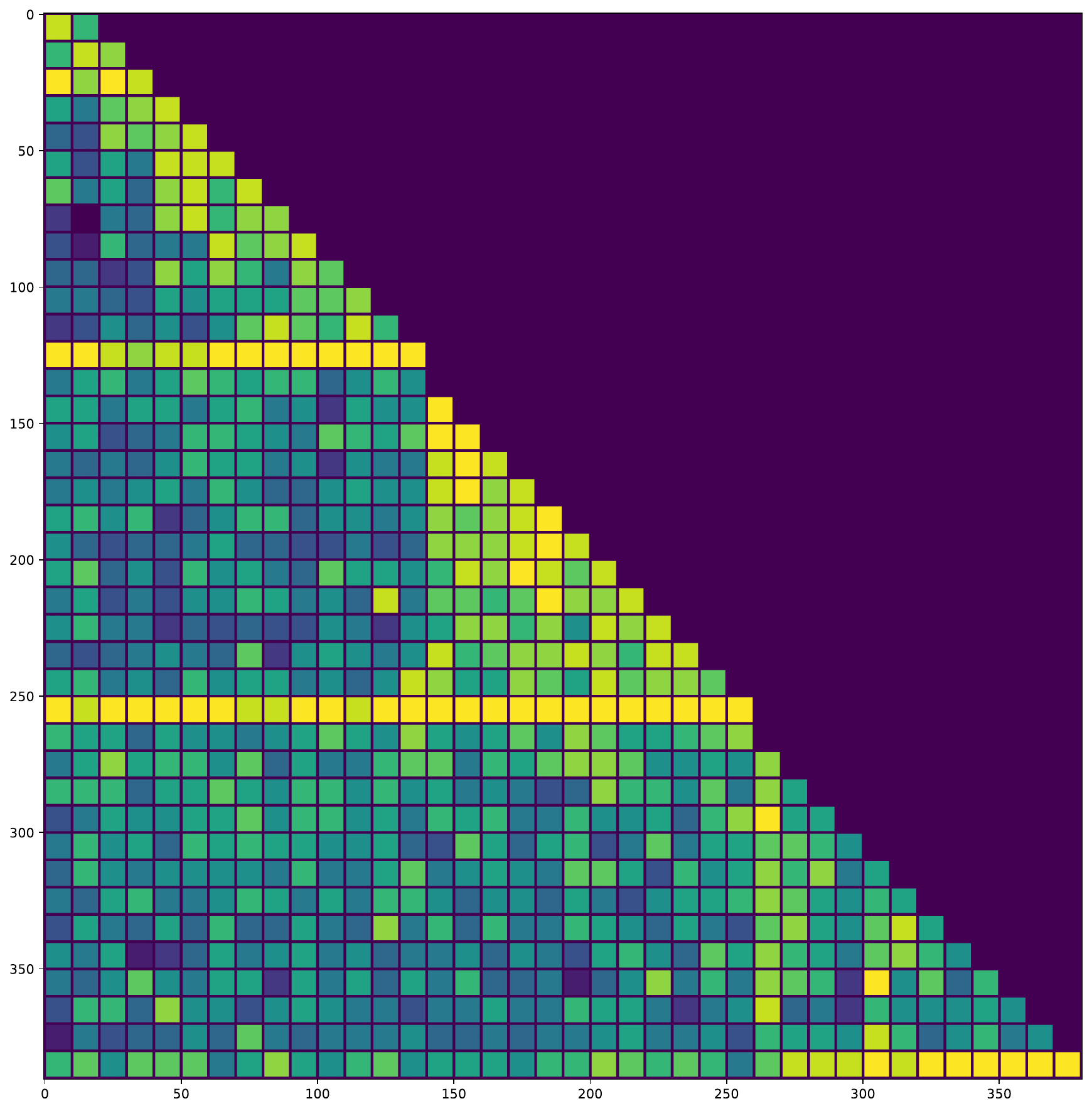}
  \caption{$l_1$-norm}
  \label{fig:sub-first}
\end{subfigure}
\begin{subfigure}{.30\textwidth}
  \centering
  \includegraphics[width=0.90\linewidth]{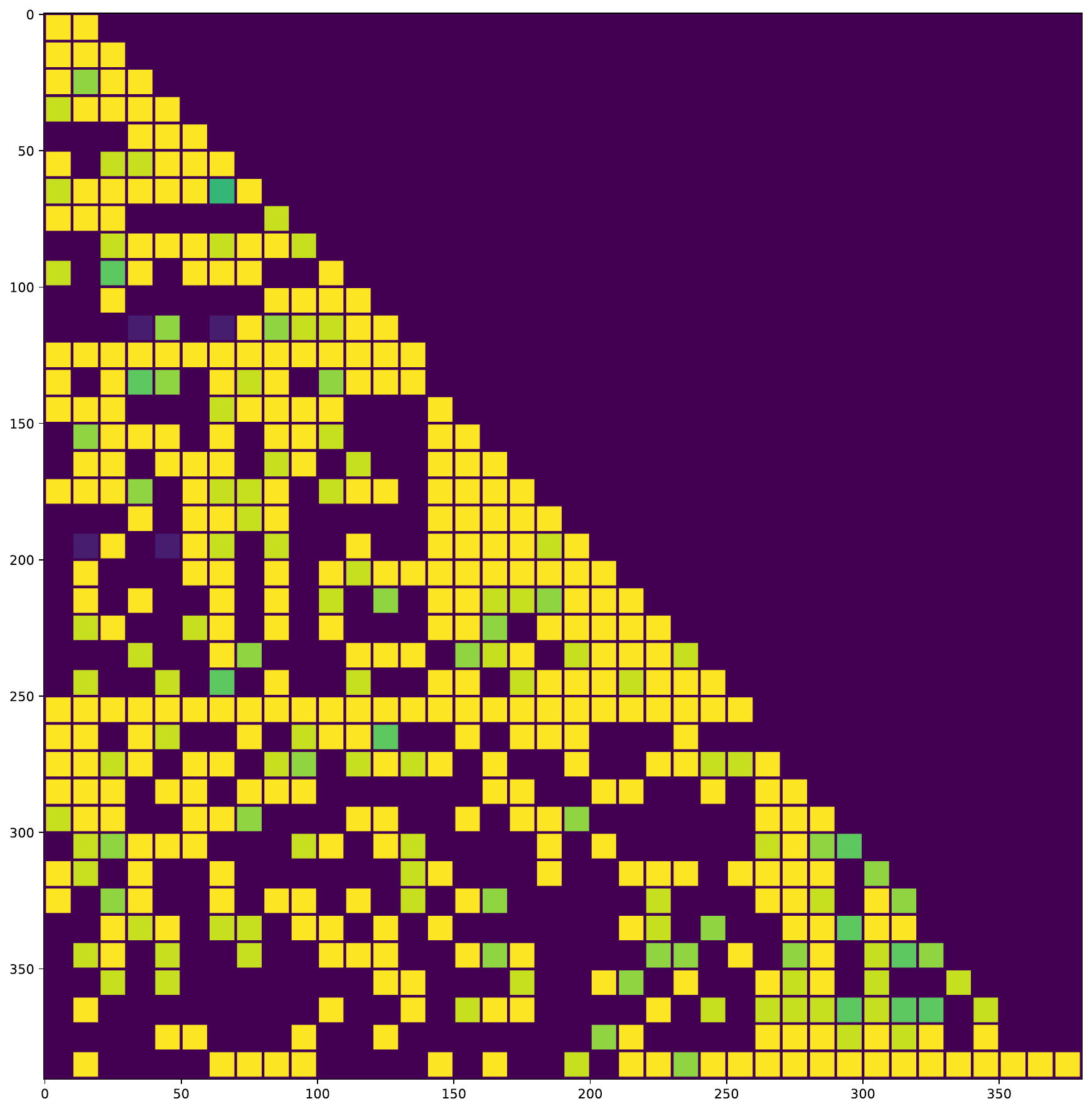}
  \caption{$l_{2,1}$-group norm}
  \label{fig:sub-second}
\end{subfigure}
\caption{Sparsified Connections in Group-wise view. DenseNet-40-K12. CIFAR100.}
\label{fig:cifar100_group}
\end{figure}

As discussed in the main paper, our method can achieve different sparsity patterns by simply switching from one norm to another for the regularization term.
An implementation example with TensorFlow is given in Listing~\ref{lst:reg}.
Figures~\ref{fig:cifar100_channel} and~\ref{fig:cifar100_group} illustrate the learned sparse structures on CIFAR100/DenseNet40-K12.
The figures clearly show that the $l_1$-norm acts on individual channels, 
and the removed channels are irregularly spread. 
On the contrary, the $l_{2,1}$-norm tends to either preserve an entire group or remove all the channels in a group. 
Table~\ref{tab:densenet40_cifar100} shows the sparsity rates and classification performances for the proposed method (DS).
The experiments with $l_{1}$-norm and $l_{2,1}$-norm exhibit similar channel sparsity ratios, 
but their group sparsity ratios are notably different. 
In this experiment, we intentionally omitted weight decay on $\alpha_i$ and $\beta$ to isolate the effects of the $l_1$- and $l_{2,1}$-norms.

\begin{table*}[!h]
\centering
\begin{small}
\begin{tabular}{*1c*2c*1c*2c*1c*2c*1c*2c*1c*2c}
\toprule
\multirow{2}{*}{Model}  &  \multicolumn{2}{c}{Channel Sparsity(\%)} & {} &  \multicolumn{2}{c}{Group Sparsity(\%)} & {} &  \multicolumn{2}{c}{Top-1 Error(\%)}  & {} & \multicolumn{2}{c}{FLOPs}  & {} & \multicolumn{2}{c}{Params} \\ \cmidrule{2-3}\cmidrule{5-6}\cmidrule{8-9}\cmidrule{11-12} \cmidrule{14-15}
{}                      &  Avg.            & Std.                   & {} &  Avg.         & Std.                    & {} &  Avg.                &  Std.          & {} & Avg.                 & Std. & {} & Avg.      & Std. \\
\midrule
Base                    &  0.0             & 0.0            & {} &  0.0         & 0.0              & {} & $25.90$   & $0.38$  & {} & $ 5.7\times10^{8} $   & 0.0 & {} & $ 1.1\times10^{6} $  & $ 0.0 $                                         \\
\midrule
DS-$l_1$                &  40.8            & 0.5            & {} & $0.1$         & 0.1             & {} & $26.07$   & $0.25$  & {} & $ 67.0\%$  & $ 0.7\% $ & {} & $62.8\%$   & $0.5\%$                               \\ 
DS-$l_{2,1}$            &  41.3            & 1.0            & {} & $39.0$        & 1.1             & {} & $26.72$   & $0.12$  & {} & $ 70.7\%$  & $ 1.3\% $ & {} & $62.4\%$   & $0.9\%$                               \\ 
\bottomrule
\end{tabular}
\end{small}
\caption{Performance comparison of $l_1$- and $l_{2,1}$-norm on CIFAR100, DenseNet-40-K12.}
\label{tab:densenet40_cifar100}
\end{table*}

\begin{listing}[h]%
\caption{Example code for Sparsity Regularization}%
\label{lst:reg}%
\begin{lstlisting}[language=Python]
def safe_l2_norm(tensor, axis=None, keepdims=None, name=None):
    @tf.custom_gradient
    def norm(x):
        y = tf.norm(x, 2, axis, keepdims, name)

        def grad(dy):
            ex_dy = tf.expand_dims(dy, axis) if axis else dy
            ex_y = tf.expand_dims(y, axis)  if axis else y
            # for numerical stability, add a small constant
            return ex_dy * (x/ (ex_y + 1e-19))

        return y, grad

    return norm(tensor)

alpha_mag = tf.nn.relu(abs_alpha - sig_beta*alpha_l1)
alpha = tf.math.sign(alpha) * alpha_mag

if norm == 'l1': #l_{1}-norm
    reg = tf.reduce_sum(alpha_mag)      
elif norm == 'group': #l_{2,1}-group norm
    # Create a group of 12 channels for DenseNet-40-K12.
    alpha_mags = tf.reshape(alpha, shape=[len(channels), -1])
    reg = tf.reduce_sum(safe_l2_norm(alpha_mags, axis=-1))
\end{lstlisting}
\end{listing}

\subsubsection{Knowledge Distillation.}

Table~\ref{tab:resnet164_kd_comp_cifar100} is a comparison table for CIFAR100/ResNet-164, omitted in the main paper due to the page limit.
Tables~\ref{tab:resnet164_kd_cifar10}, \ref{tab:resnet164_kd_cifar100} and \ref{tab:resnet50_kd_imagenet} present additional experimental results on CIFAR$10/100$ and ImageNet$2012$~\cite{Deng2009ImageNet} 
for our proposed approach. 
For CIFAR$10/100$, the reported values in the tables are averages of five repetitions. 
In the case of ImageNet$2012$, each experiment was conducted once. 
To achieve varying degrees of sparsity, we experimented with different values of $\lambda$. 
In the tables, we ranged the values of $\lambda$
from $0.000010$ to $0.000125$ for CIFAR10,  
from $0.000100$ to $0.000225$ for CIFAR100
and 
from $0.000175$ to $0.000300$ for ImageNet2012.
Tables~\ref{tab:densenet_kd_cifar10} and \ref{tab:densenet_kd_cifar100} present additional experimental results on CIFAR$10/100$ with DenseNet-40-K12.
In the tables, we ranged the values of $\lambda$ 
from $0.000010$ to $0.000125$ for CIFAR10, 
and from $0.000050$ to $0.000400$ for CIFAR100.

\begin{table}[!h]
\centering
\begin{small}
\begin{tabular}{*1c*1c*1c*1c}
\toprule
   Method             &   Top-1 Error~/~BL(\%)  &FLOPs(\%)   &  Params(\%)  \\
\midrule

SSS\tablefootnote{\cite{Huang2018ECCV}} & 24.42~/~23.31 &  55.33 & 86.75 \\
Hinge\tablefootnote{\cite{Li2020}}       & 23.12~/~23.22 &  55.32 & 76.57 \\
\midrule
\multirow{2}{*}{Ours}                         & 21.32~/~22.82 &  47.74 & 75.28 \\
{}                                            & 22.23~/~22.82 &  31.67 & 58.55 \\
\bottomrule
\end{tabular}
\end{small}
\caption{Knowledge Distillation. CIFAR100. ResNet-164.}
\label{tab:resnet164_kd_comp_cifar100}
\end{table}

\begin{table*}[!h]
\centering
\begin{small}
\begin{tabular}{*1c*2c*1c*2c*1c*2c*1c*2c*1c*2c}
\toprule
\multirow{2}{*}{Model}  &  \multicolumn{2}{c}{Channel Sparsity(\%)} & {} &  \multicolumn{2}{c}{Layer Sparsity(\%)} & {} &  \multicolumn{2}{c}{Top-1 Error(\%)} & {} & \multicolumn{2}{c}{FLOPs}  & {}  &  \multicolumn{2}{c}{Parmas}   \\\cmidrule{2-3}\cmidrule{5-6}\cmidrule{8-9}\cmidrule{11-12} \cmidrule{14-15}
{}                      &  Avg.            & Std.                   & {} &  Avg.         & Std.                    & {} &  Avg.         & Std.                 & {} &  Avg.               &  Std.      & {} & Avg.                 & Std. \\
\midrule
Base                    &  0.0             & 0.0                    & {} &  0.0          & 0.0                     & {} &  4.72         & 0.10                 & {} & $ 5.0\times10^{8} $  & 0.0 & {}  &  $1.7\times10^{6}$  & 0.0     \\
\midrule
\multirow{6}{*}{DS}     &  43.6            & 0.3                    & {} &  1.1          & 0.9                     & {} &  4.55         & 0.12                 & {} & $ 70.0\%$           & $ 1.7\% $  & {} &  $83.9\%$           & $0.4\%$     \\
{}                      &  53.1            & 0.3                    & {} &  5.3          & 2.6                     & {} &  4.70         & 0.13                 & {} & $ 58.7\%$            & $ 1.1\% $ & {} &  $75.5\%$           & $0.2\%$ \\
{}                      &  60.7            & 0.3                    & {} &  15.7         & 2.2                     & {} &  4.84         & 0.07                 & {} & $ 47.2\%$            & $ 0.7\% $ & {} &  $66.9\%$           & $0.7\%$  \\
{}                      &  66.7            & 0.3                    & {} &  28.2         & 2.7                     & {} &  4.99         & 0.11                 & {} & $ 39.1\%$            & $ 0.4\% $ & {} &  $59.6\%$           & $0.9\%$  \\
{}                      &  70.5            & 0.8                    & {} &  36.4         & 8.3                     & {} &  5.27         & 0.22                 & {} & $ 33.9\%$            & $ 1.6\% $ & {} &  $53.7\%$           & $0.4\%$  \\
{}                      &  73.9            & 0.3                    & {} &  40.4         & 2.7                     & {} &  5.37         & 0.22                 & {} & $ 29.5\%$            & $ 0.7\% $ & {} &  $48.0\%$           & $0.5\%$ \\
\bottomrule
\end{tabular}
\end{small}
\caption{Knowledge Distillation. Performance on CIFAR10. ResNet-164.}
\label{tab:resnet164_kd_cifar10}
\end{table*}

\begin{table*}[!h]
\centering
\begin{small}
\begin{tabular}{*1c*2c*1c*2c*1c*2c*1c*2c*1c*2c}
\toprule
\multirow{2}{*}{Model}  &  \multicolumn{2}{c}{Channel Sparsity(\%)} & {} &  \multicolumn{2}{c}{Layer Sparsity(\%)} & {} &  \multicolumn{2}{c}{Top-1 Error(\%)} & {} & \multicolumn{2}{c}{FLOPs}  & {} &  \multicolumn{2}{c}{Parmas}  \\\cmidrule{2-3}\cmidrule{5-6}\cmidrule{8-9}\cmidrule{11-12} \cmidrule{14-15}
{}                      &  Avg.            & Std.                   & {} &  Avg.         & Std.                    & {} &  Avg.         & Std.                 & {} &  Avg.               &  Std.     & {} & Avg.                 & Std. \\
\midrule
Base                    &  0.0             & 0.0                    & {} &  0.0          & 0.0                     & {} &  22.82        & 0.33                 & {} & $ 5.0\times10^{8} $ & 0.0       & {} &  $1.7\times10^{6}$  & 0.0     \\
\midrule
\multirow{6}{*}{DS}     &  55.3            & 0.4                    & {} &  26.5         & 1.2                     & {} &  21.32        & 0.19                 & {} & $ 47.7\%$           & $ 0.6\% $ & {} &  $75.3\%$           & $0.6\%$      \\
{}                      &  58.4            & 0.4                    & {} &  26.9         & 4.0                     & {} &  21.42        & 0.18                 & {} & $ 45.2\%$           & $ 0.7\% $ & {} &  $72.8\%$           & $0.2\%$  \\
{}                      &  62.1            & 0.6                    & {} &  33.3         & 5.5                     & {} &  21.94        & 0.17                 & {} & $ 41.1\%$           & $ 1.6\% $ & {} &  $68.5\%$           & $0.5\%$  \\
{}                      &  65.3            & 0.2                    & {} &  39.1         & 3.0                     & {} &  22.12        & 0.16                 & {} & $ 37.1\%$           & $ 0.7\% $ & {} &  $65.1\%$           & $0.4\%$ \\
{}                      &  67.9            & 0.3                    & {} &  42.3         & 1.2                     & {} &  22.42        & 0.23                 & {} & $ 34.0\%$           & $ 0.5\% $ & {} &  $61.6\%$           & $0.3\%$  \\
{}                      &  69.9            & 0.4                    & {} &  46.8         & 1.9                     & {} &  22.23        & 0.18                 & {} & $ 31.7\%$           & $ 0.9\% $ & {} &  $58.6\%$           & $0.5\%$  \\
\bottomrule
\end{tabular}
\end{small}
\caption{Knowledge Distillation. Performance on CIFAR100. ResNet-164.}
\label{tab:resnet164_kd_cifar100}
\end{table*}

\begin{table}[!h]
\centering
\begin{small}
\begin{tabular}{*1c*1c*1c*1c*1c*1c}
\toprule
   Model                & Channel Sparsity (\%) &  Layer Sparsity (\%) &    Top-1 Error(\%)  & FLOPs            &  Params             \\
\midrule      
Base                    &  0.0                  &   0.0                &    24.77            & $8.2\times10^{9}$ &  $2.6\times10^{7}$  \\     

\midrule                
\multirow{6}{*}{DS}   &  59.8                 &   0.0                &    24.73            &  60.1\%           &  65.9\%                   \\
{}                      &  62.8                 &   0.0                &    25.17            &  55.7\%           &  62.0\%                   \\
{}                      &  65.1                 &   0.0                &    25.58            &  53.7\%           &  58.9\%                   \\
{}                      &  68.2                 &   12.2               &    26.42            &  46.4\%           &  54.9\%                   \\
{}                      &  70.2                 &   12.2               &    26.88            &  44.1\%           &  51.5\%                   \\
{}                      &  72.7                 &   24.5               &    27.71            &  41.3\%           &  47.7\%             \\
\bottomrule
\end{tabular}
\end{small}
\caption{Knowledge Distillation. ImageNet2012. ResNet-50.}
\label{tab:resnet50_kd_imagenet}
\end{table}

\begin{table*}[!h]
\centering
\begin{small}
\begin{tabular}{*1c*2c*1c*2c*1c*2c*1c*2c}
\toprule
\multirow{2}{*}{Model}  &  \multicolumn{2}{c}{Channel Sparsity(\%)} & {} &  \multicolumn{2}{c}{Top-1 Error(\%)} & {} & \multicolumn{2}{c}{FLOPs}        & {} &  \multicolumn{2}{c}{Parmas}     \\\cmidrule{2-3}\cmidrule{5-6}\cmidrule{8-9}\cmidrule{11-12}
{}                      &  Avg.            & Std.                   & {} &  Avg.         & Std.                 & {} &  Avg.               &  Std.      & {} & Avg.                 & Std. \\
\midrule
Base                    &  0.0             & 0.0                    & {} &  5.32         & 0.24                 & {} & $ 5.7\times10^{8} $  & 0.0       & {} &  $1.1\times10^{6}$  & 0.0     \\
\midrule
\multirow{6}{*}{DS}     &  41.3            & 0.5                    & {} &  5.05         & 0.21                 & {} & $ 63.3\%$            & $ 0.9\% $ & {} &  $63.7\%$           & $0.6\%$ \\
{}                      &  53.9            & 0.3                    & {} &  4.96         & 0.11                 & {} & $ 52.3\%$            & $ 0.8\% $ & {} &  $51.3\%$           & $0.3\%$ \\
{}                      &  62.5            & 0.3                    & {} &  4.91         & 0.05                 & {} & $ 45.1\%$            & $ 0.7\% $ & {} &  $42.8\%$           & $0.4\%$\\
{}                      &  68.1            & 0.4                    & {} &  5.20         & 0.04                 & {} & $ 40.4\%$            & $ 0.3\% $ & {} &  $37.2\%$           & $0.3\%$  \\
{}                      &  71.9            & 0.2                    & {} &  5.32         & 0.18                 & {} & $ 36.8\%$            & $ 0.5\% $ & {} &  $33.3\%$           & $0.2\%$  \\
{}                      &  74.8            & 0.4                    & {} &  5.35         & 0.30                 & {} & $ 34.5\%$            & $ 0.6\% $ & {} &  $30.4\%$           & $0.4\%$ \\
\bottomrule
\end{tabular}
\end{small}
\caption{Knowledge Distillation. Performance on CIFAR10. DenseNet-40-12K.}
\label{tab:densenet_kd_cifar10}
\end{table*}

\begin{table*}[!h]
\centering
\begin{small}
\begin{tabular}{*1c*2c*1c*2c*1c*2c*1c*2c}
\toprule
\multirow{2}{*}{Model}  &  \multicolumn{2}{c}{Channel Sparsity(\%)} & {} &  \multicolumn{2}{c}{Top-1 Error(\%)} & {} & \multicolumn{2}{c}{FLOPs}        & {} &  \multicolumn{2}{c}{Parmas}   \\\cmidrule{2-3}\cmidrule{5-6}\cmidrule{8-9}\cmidrule{11-12}
{}                      &  Avg.            & Std.                   & {} &  Avg.         & Std.                 & {} &  Avg.                &  Std.     & {} & Avg.                 & Std. \\
\midrule
Base                    &  0.0             & 0.0                    & {} &  25.73        & 0.49                 & {} & $ 5.7\times10^{8} $  & 0.0       & {} &  $1.1\times10^{6}$  & 0.0      \\
\midrule
\multirow{6}{*}{DS}     &  40.3            & 0.3                    & {} &  23.14        & 0.18                 & {} & $ 64.9\%$            & $ 0.6\% $ & {} &  $63.4\%$           & $0.3\%$  \\
{}                      &  56.1            & 0.4                    & {} &  23.40        & 0.09                 & {} & $ 53.0\%$            & $ 0.9\% $ & {} &  $48.2\%$           & $0.4\%$  \\
{}                      &  68.5            & 0.2                    & {} &  23.76        & 0.37                 & {} & $ 43.6\%$            & $ 0.3\% $ & {} &  $36.2\%$           & $0.2\%$  \\
{}                      &  73.4            & 0.3                    & {} &  24.09        & 0.17                 & {} & $ 38.7\%$            & $ 0.5\% $ & {} &  $31.4\%$           & $0.3\%$  \\
{}                      &  78.3            & 0.1                    & {} &  24.33        & 0.27                 & {} & $ 34.0\%$            & $ 0.5\% $ & {} &  $26.4\%$           & $0.2\%$ \\
{}                      &  82.4            & 0.2                    & {} &  25.01        & 0.32                 & {} & $ 29.7\%$            & $ 0.4\% $ & {} &  $22.2\%$           & $0.2\%$  \\
\bottomrule
\end{tabular}
\end{small}
\caption{Knowledge Distillation. Performance on CIFAR100. DenseNet-40-12K.}
\label{tab:densenet_kd_cifar100}
\end{table*}

\subsection{Rectified Gradient Flow and Discovering Neural Wiring}

\subsubsection{Code.}
Listing~\ref{lst:rgf} shows the TensorFlow implementation of the rectified gradient flow.
The learning mechanism can be toggled on and off by simply switching from \emph{tf.nn.relu} to \emph{rgf\_relu}, or vice versa.

\begin{listing}[tb]%
\caption{Example code for Rectified Gradient Flow}%
\label{lst:rgf}%
\begin{lstlisting}[language=Python]
@tf.custom_gradient
def rgf_relu(x):
    o1 = tf.nn.relu(x) # forward-pass
    o2 = tf.keras.activations.elu(x, alpha=0.1) # backward-pass
    
    def grad(dy): # the gradient of elu is used in backward-pass
        return tf.gradients(o2, [x], grad_ys=[dy])

    return o1, grad     
    
abs_alpha = tf.math.abs(alpha)        
alpha_l1 = tf.reduce_sum(abs_alpha, axis=2, keepdims=True)

if rgf is True: alpha_mag = rgf_relu(abs_alpha - sigmoid_beta*alpha_l1) 
else: alpha_mag = tf.nn.relu(abs_alpha - sigmoid_beta*alpha_l1) 
        
alpha = tf.math.sign(alpha) * alpha_mag

reg = tf.reduce_sum(tf.math.abs(alpha_mag)) #l_{1}-norm
\end{lstlisting}
\end{listing}

\subsubsection{Connection to DNW.}

Assume that an output $y$ of a neuron in a hidden layer is written as

\[
  y \left( \textbf{x}\right) = \sum_{i=1}^{n}{a_i  f_i\left(\textbf{x}; \textbf{w}_i\right)},
\] where \textbf{x} denotes an input from a preceding layer, $\textbf{w}_i$ represents the model parameters for component $f_i$,
and $a_{i}$ is an edge or an architecture parameter.
A loss function can be denoted
\[
\mathcal{L}\left( a \left(\alpha\right), f \left(\textbf{x}; \textbf{w} \right)\right).
\]
Let $a_i$ be a function of $\alpha_i$,
\[
a_{i}\left( \alpha_i\right) = sign\left(\alpha_{i}\right) \left(\abs{\alpha_{i}} - \sigma\left(\beta \right) \right)_{+},
\] which is simplified for this analysis.
The gradients can be written as
\begin{equation}\label{eqn:grad_alpha}
\frac{\partial \mathcal{L}}{\partial \alpha_i } = \frac{\partial \mathcal{L}}{\partial y} \cdot \frac{\partial a_i \left( \alpha_i \right) }{\partial \alpha_i} \cdot f_i \left(\textbf{x}, \textbf{w}_i \right),
\end{equation}

\begin{equation}\label{eqn:grad_w}
\frac{\partial \mathcal{L}}{\partial \textbf{w}_{i} } = \frac{\partial \mathcal{L}}{\partial y} \cdot  a_i\left(\alpha_i\right) \cdot \frac{\partial f_i \left(\textbf{x}; \textbf{w}_i \right) }{\partial \textbf{w}_{i}} ,
\end{equation}

\begin{equation}\label{eqn:grad_x}
\frac{\partial \mathcal{L}}{\partial \textbf{x} } = \frac{\partial \mathcal{L}}{\partial y} \cdot \left(\sum_{j\neq i}^{n}  a_j\left(\alpha_j\right) \cdot \frac{\partial f_j \left(\textbf{x}; \textbf{w}_j \right) }{\partial \textbf{x}} + a_i\left(\alpha_i\right) \cdot \frac{\partial f_i \left(\textbf{x}; \textbf{w}_i \right) }{\partial \textbf{x}}\right).
\end{equation}
If $\abs{\alpha_{i}} < \sigma\left(\beta \right)$, $\partial a_i / \partial \alpha_i$ in Eq.~(\ref{eqn:grad_alpha}) is zero and $\alpha_i$ is not updated. 
However, if $elu$~\cite{Clevert2016} is employed in the backward pass, an approximate gradient for $\alpha_i$ can be generated and $\alpha_i$ can be updated.
Since $ a_i=0$ in Eq.~(\ref{eqn:grad_w}) and~(\ref{eqn:grad_x}) when $\abs{\alpha_{i}} < \sigma\left(\beta \right)$,
they do not receive a learning signal through $a_i$ even if the rectified gradient flow is adopted.
This results in a similar learning mechanism proposed in DNW~\cite{Wortsman2019}, where the gradient flows to zeroed-out (hallucinated) edges but not through them.


\subsubsection{Experiment.}

Tables~\ref{tab:dnw_cifar10} and~\ref{tab:dnw_cifar100} provide additional experimental results on CIFAR10/100 for our proposed method (DS).
Following the practice of DNW, 
we employed MobileNetV$1\left(\times 0.25\right)$~\cite{Andrew2017} as a base model.
In DNW, the value of $k$ was chosen such that the final learned model has similar Mult-Adds as
the base model. We set the value of $\lambda$ in the same manner.
We ran each experiment $5$ times and show the average and standard deviation.

\begin{table*}[!h]
\centering
\begin{small}
\begin{tabular}{*1c*2c*1c*2c*1c*2c}
\toprule
\multirow{2}{*}{Model}   & \multicolumn{2}{c}{Top-1 Error(\%)} & {} & \multicolumn{2}{c}{Mult-Adds}         & {} &  \multicolumn{2}{c}{Parmas}               \\\cmidrule{2-3}\cmidrule{5-6}\cmidrule{8-9}
{}                       &  Avg.           & Std.              & {} &  Avg.                &  Std.             & {} & Avg.                 & Std. \\
\midrule
MobileNetV1($\times0.25$)& 13.44           & 0.24              & {} & $ 3.3\times10^{6} $   & 0.0           & {} & $2.2\times10^{5}$   & 0.0                \\
\midrule
{}                       &\multicolumn{8}{c}{DNW($\times0.225$)} \\
\midrule
No Update Rule&13.86     & 0.27                                & {} & $ 4.5\times10^{6} $   & $ 3.7\times10^{4} $ & {} & $2.2\times10^{5}$   & $3.7\times10^{1}$ \\
With Update Rule  &10.30 & 0.20                                & {} & $ 3.1\times10^{6} $   & $ 4.6\times10^{4} $ & {} & $1.8\times10^{5}$   & $6.7\times10^{1}$ \\
\midrule
$ \lambda \times 10^{-3}$&\multicolumn{8}{c}{Proximal Gradient with $l_{1}$-norm} \\
\midrule
1.875  & 11.64       & 0.41                                   & {} & $ 3.7\times10^{6} $   & $ 1.3\times10^{5} $  & {} & $2.4\times10^{4}$   & $9.0\times10^{2}$  \\
1.950  & 11.95       & 0.42                                   & {} & $ 3.7\times10^{6} $   & $ 2.6\times10^{5} $  & {} & $2.3\times10^{4}$   & $1.5\times10^{3}$  \\
2.225  & 12.17       & 0.44                                   & {} & $ 3.3\times10^{6} $   & $ 1.7\times10^{5} $  & {} & $2.1\times10^{4}$   & $9.4\times10^{2}$  \\
2.325  & 12.50       & 0.28                                   & {} & $ 3.2\times10^{6} $   & $ 1.1\times10^{5} $  & {} & $1.9\times10^{4}$   & $5.7\times10^{2}$  \\
2.400  & 12.66       & 0.38                                   & {} & $ 3.0\times10^{6} $   & $ 6.4\times10^{4} $  & {} & $1.8\times10^{4}$   & $1.1\times10^{3}$  \\
\midrule
$ \lambda \times 10^{-3}$&\multicolumn{8}{c}{Proximal Gradient with $l_{1,2}$-norm} \\
\midrule
1.250  & 12.15       & 1.17                                  & {} & $ 3.7\times10^{6} $   & $ 1.2\times10^{5} $   & {} & $1.0\times10^{5}$   & $1.3\times10^{4}$  \\
1.375  & 13.14       & 0.42                                  & {} & $ 3.6\times10^{6} $   & $ 2.1\times10^{5} $   & {} & $9.6\times10^{4}$   & $1.2\times10^{4}$  \\
1.500  & 13.62       & 0.56                                  & {} & $ 3.4\times10^{6} $   & $ 8.6\times10^{4} $   & {} & $9.6\times10^{4}$   & $1.6\times10^{4}$  \\
1.625  & 14.12       & 0.62                                  & {} & $ 3.3\times10^{6} $   & $ 1.6\times10^{5} $   & {} & $8.7\times10^{4}$   & $8.6\times10^{3}$  \\
1.750  & 15.09       & 1.21                                  & {} & $ 3.2\times10^{6} $   & $ 2.0\times10^{5} $   & {} & $8.7\times10^{4}$   & $1.5\times10^{4}$  \\
\midrule
$ \lambda \times 10^{-5}$&\multicolumn{8}{c}{DS-No Rectified Gradient Flow} \\
\midrule
1.125    & 10.20         & 0.08                             & {} & $ 3.6\times10^{6} $   & $ 4.9\times10^{4} $    & {} & $6.9\times10^{4}$   & $8.9\times10^{2}$      \\
1.375    & 10.55         & 0.23                             & {} & $ 3.4\times10^{6} $   & $ 4.5\times10^{4} $    & {} & $6.1\times10^{4}$   & $5.7\times10^{2}$       \\
1.625    & 10.96         & 0.18                             & {} & $ 3.2\times10^{6} $   & $ 5.3\times10^{4} $    & {} & $5.6\times10^{4}$   & $1.7\times10^{3}$      \\
1.875    & 11.06         & 0.45                             & {} & $ 3.1\times10^{6} $   & $ 3.7\times10^{4} $    & {} & $5.1\times10^{4}$   & $2.4\times10^{2}$      \\
2.000    & 11.05         & 0.25                             & {} & $ 3.0\times10^{6} $   & $ 3.4\times10^{4} $    & {} & $4.9\times10^{4}$   & $8.0\times10^{2}$       \\
\midrule
$ \lambda \times 10^{-5}$&\multicolumn{8}{c}{DS-Rectified Gradient Flow} \\
\midrule
6.0                    & 9.04         & 0.10                 & {} & $ 3.5\times10^{6} $   &$ 4.8\times10^{4} $    & {} & $5.3\times10^{4}$   & $7.0\times10^{2}$         \\
6.5                    & 9.28         & 0.38                 & {} & $ 3.5\times10^{6} $   &$ 7.8\times10^{4} $    & {} & $5.0\times10^{4}$   & $9.8\times10^{2}$       \\
7.0                    & 9.36         & 0.27                 & {} & $ 3.3\times10^{6} $   &$ 6.7\times10^{4} $    & {} & $4.7\times10^{4}$   & $8.4\times10^{2}$        \\
7.5                    & 9.32         & 0.19                 & {} & $ 3.3\times10^{6} $   &$ 8.5\times10^{4} $    & {} & $4.5\times10^{4}$   & $3.2\times10^{2}$       \\
8.0                    & 9.66         & 0.07                 & {} & $ 3.1\times10^{6} $   &$ 6.0\times10^{4} $    & {} & $4.2\times10^{4}$   & $9.0\times10^{2}$       \\
\bottomrule
\end{tabular}
\end{small}
\caption{Performance on CIFAR10, Discovering Neural Wiring.}
\label{tab:dnw_cifar10}
\end{table*}

\begin{table*}[!h]
\centering
\begin{small}
\begin{tabular}{*1c*2c*1c*2c*1c*2c}
\toprule
\multirow{2}{*}{Model}  & \multicolumn{2}{c}{Top-1 Error(\%)} & {} & \multicolumn{2}{c}{Mult-Adds}              & {} &  \multicolumn{2}{c}{Parmas}               \\\cmidrule{2-3}\cmidrule{5-6}\cmidrule{8-9}
{}                      &  Avg.           & Std.              & {} &  Avg.                &  Std.               & {} & Avg.                 & Std. \\
\midrule
MobileNetV1($\times0.25$)& 43.78          & 0.54              & {} & $ 3.4\times10^{6} $  & 0.0                 & {} & $2.4\times10^{5}$   & 0.0                \\
\midrule
{}                       &\multicolumn{8}{c}{DNW($\times0.225$)} \\
\midrule
No Update Rule & 40.50           & 0.30                       & {} & $ 4.6\times10^{6} $   & $ 3.6\times10^{4} $ & {} & $3.1\times10^{5}$   & $3.8\times10^{1}$           \\
With Update Rule      &34.18            & 0.37                & {} & $ 3.3\times10^{6} $   & $ 4.7\times10^{4} $ & {} & $2.6\times10^{5}$   & $5.0\times10^{2}$       \\
\midrule
$ \lambda \times 10^{-3}$&\multicolumn{8}{c}{Proximal Gradient with $l_{1}$-norm} \\
\midrule
2.1   & 35.27       & 0.65                                    & {} & $ 3.9\times10^{6} $   & $ 2.6\times10^{5} $ & {} & $1.3\times10^{5}$   & $4.5\times10^{3}$  \\
2.2   & 36.00       & 0.70                                    & {} & $ 3.6\times10^{6} $   & $ 9.5\times10^{4} $ & {} & $1.2\times10^{5}$   & $3.9\times10^{3}$ \\
2.3   & 35.21       & 0.32                                    & {} & $ 3.6\times10^{6} $   & $ 1.4\times10^{5} $ & {} & $1.2\times10^{5}$   & $4.1\times10^{3}$  \\
2.4   & 36.57       & 0.53                                    & {} & $ 3.4\times10^{6} $   & $ 1.9\times10^{5} $ & {} & $1.2\times10^{5}$   & $4.3\times10^{3}$  \\
2.5   & 36.81       & 0.41                                    & {} & $ 3.2\times10^{6} $   & $ 2.0\times10^{5} $ & {} & $1.1\times10^{5}$   & $3.5\times10^{3}$  \\
\midrule
$ \lambda \times 10^{-3}$&\multicolumn{8}{c}{Proximal Gradient with $l_{1,2}$-norm} \\
\midrule

1.00  & 35.97       & 0.59                                    & {} & $ 4.4\times10^{6} $   & $ 1.2\times10^{5} $ & {} & $3.0\times10^{5}$   & $7.7\times10^{3}$  \\
1.25  & 37.52       & 0.84                                    & {} & $ 3.9\times10^{6} $   & $ 1.2\times10^{5} $ & {} & $2.7\times10^{5}$   & $6.3\times10^{3}$ \\
1.50  & 38.45       & 0.29                                    & {} & $ 3.7\times10^{6} $   & $ 1.1\times10^{5} $ & {} & $2.6\times10^{5}$   & $5.5\times10^{3}$  \\
1.75  & 39.63       & 0.54                                    & {} & $ 3.5\times10^{6} $   & $ 1.4\times10^{5} $ & {} & $2.5\times10^{5}$   & $1.1\times10^{4}$  \\
2.00  & 41.35       & 0.86                                    & {} & $ 3.3\times10^{6} $   & $ 1.6\times10^{5} $ & {} & $2.5\times10^{5}$   & $9.7\times10^{3}$ \\
\midrule
$ \lambda \times 10^{-5}$&\multicolumn{8}{c}{DS-No Rectified Gradient Flow} \\
\midrule
2.00   & 35.51         & 0.62                                 & {} & $ 3.7\times10^{6} $   & $ 4.8\times10^{4} $ & {} & $1.8\times10^{5}$   & $5.4\times10^{2}$       \\
2.25   & 35.26         & 0.50                                 & {} & $ 3.5\times10^{6} $   & $ 4.6\times10^{4} $ & {} & $1.7\times10^{5}$   & $7.8\times10^{2}$       \\
2.50   & 35.68         & 0.47                                 & {} & $ 3.4\times10^{6} $   & $ 4.3\times10^{4} $ & {} & $1.7\times10^{5}$   & $7.5\times10^{2}$       \\
2.75   & 35.32         & 0.28                                 & {} & $ 3.3\times10^{6} $   & $ 2.5\times10^{4} $ & {} & $1.6\times10^{5}$   & $5.2\times10^{2}$      \\
3.00   & 35.77         & 0.36                                 & {} & $ 3.2\times10^{6} $   & $ 2.3\times10^{4} $ & {} & $1.6\times10^{5}$   & $1.0\times10^{3}$       \\

\midrule
$ \lambda \times 10^{-4}$&\multicolumn{8}{c}{DS-Rectified Gradient Flow} \\
\midrule
0.850   & 32.16        & 0.18                                 & {} & $ 3.8\times10^{6} $   &$ 1.2\times10^{5} $  & {} & $2.0\times10^{5}$   & $3.0\times10^{3}$          \\
0.925   & 32.84        & 0.44                                 & {} & $ 3.6\times10^{6} $   &$ 8.6\times10^{4} $  & {} & $1.9\times10^{5}$   & $4.8\times10^{3}$         \\
1.000   & 32.78        & 0.83                                 & {} & $ 3.5\times10^{6} $   &$ 1.2\times10^{5} $  & {} & $1.8\times10^{5}$   & $3.3\times10^{3}$         \\
1.125   & 32.92        & 0.60                                 & {} & $ 3.3\times10^{6} $   &$ 6.8\times10^{4} $  & {} & $1.6\times10^{5}$   & $4.6\times10^{3}$          \\
1.250   & 33.80        & 0.58                                 & {} & $ 3.1\times10^{6} $   &$ 8.9\times10^{4} $  & {} & $1.6\times10^{5}$   & $4.4\times10^{3}$\\
\bottomrule
\end{tabular}
\end{small}
\caption{Performance on CIFAR100, Discovering Neural Wiring.}
\label{tab:dnw_cifar100}
\end{table*}

\subsection{Learning Sparse Affinity Matrix}

Part of this section will be revised with appropriate citations upon official publication of the paper.

\subsubsection{Traffic Speed Data.}

\begin{figure}[!t]
\begin{center}
\centerline{\includegraphics[width=0.275\columnwidth]{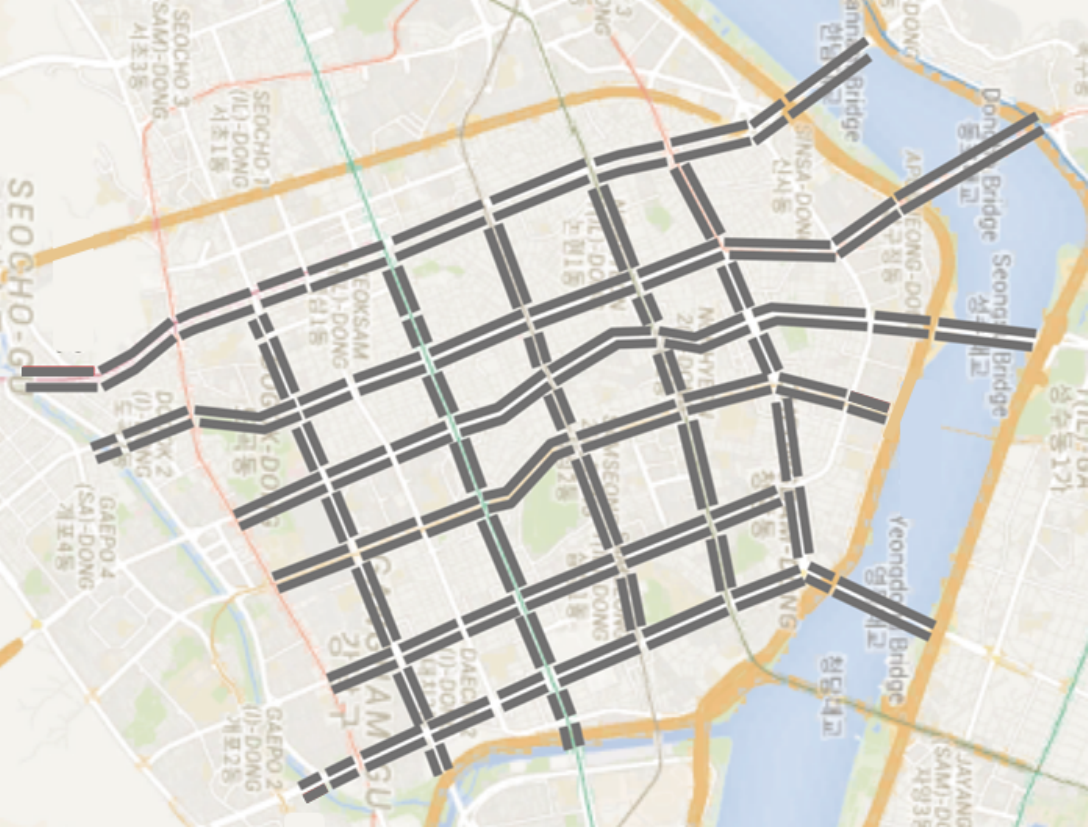}}
\caption{The gray lines represent the $170$ road links where the experimental data were collected.}
\label{fig:map}
\end{center}
\end{figure}

\begin{table*}[!t]
\centering
\begin{small}
\begin{tabular}{*6c}
\toprule
\multirow{2}{*}{Area}  & \# Road & Collection      & Collection       & \multirow{2}{*}{Time Interval}  & Time Horizon   \\
{}                     & Segments&  Method         &  Period          &                            {}   & for Prediction \\\cmidrule{1-6}
See Fig.~\ref{fig:map} &  $170$    &  Taxi with GPS  & Jan.-Oct.  &  15 min.                        & 1 (15 min.)  \\
\bottomrule
\end{tabular}
\end{small}
\caption{Summary of Experimental Data}
\label{tab:traffic_data}
\end{table*}

A summary of the data used in the main paper is shown in Table~\ref{tab:traffic_data}.
There were approximately $70,000$ probe taxis operating in the metropolitan area where the data were collected.
As the taxis operated in three shifts,
more than $20,000$ probes on average ran at any given point in time.
The traffic speed data covers $4,663$ links in the area, which includes major arterials.
Speed data from probes in the metropolitan area were aggregated every $5$ minutes for each road segment;
if no probe passed a link during a $5$-minute period,
the speed was estimated based on speeds in the previous time periods or from the same time periods in the past.
After preprocessing, the raw data were once again aggregated in $15$-minute periods to maximize forecasting utility.
A $15$-minute period is the standard on which the highway capacity manual is based.
As no information was available regarding probe counts, which contribute to averaging speed data,
three speeds for a $5$-minute period were averaged without weighting.
Among the $4,663$ links for which speed data were available,
we selected $170$ links in the subarea (see Fig.~\ref{fig:map}).
We expected that the speed data collected in this region would be free from missing observations
as it is the busiest region in the metropolitan area, and most taxi drivers congregate in this region to find passengers.
Following the generic conventions of machine learning,
we used data from the previous eight months to train the proposed model and reserved the data for the latter two months to test the trained model.

\subsubsection{GCN Prediction Model.}
\begin{figure}[!htbp]
\centering
\includegraphics[width=.50\columnwidth]{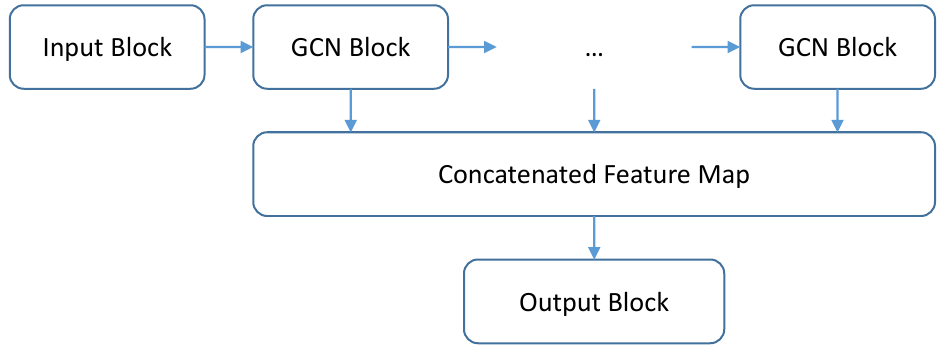}
\caption{
Graph convolutional neural network (GCN) structure.
The GCN in our experiments consists of $5$ GCN blocks.
The output feature maps of GCN blocks are concatenated and then fed to an output block.
Input and Output blocks consist of 3 fully connected layers.
}
\label{fig:gcn_exp_02}
\end{figure}

Figure~\ref{fig:gcn_exp_02} illustrates our GCN implementation.
The outputs of hidden GCN blocks are concatenated and fed to an output block.
This was motivated by the dense connections of DenseNet~\cite{Huang2017}.
The structures of the hidden GCN blocks are illustrated in the main paper.
We trained the model for $500$ epochs using Adam~\cite{Kingma2015}
with the defaulting parameter setting of TensorFlow.
The initial learning rate was set to $0.0005$ and multiplied by $0.5$ at epochs $400$ and $450$.

A training loss is defined by the mean relative error (MRE)
\[
\frac{1}{N} \sum^{N}_{i=1}\frac{| \tilde{y}_i - y_i|}{y_i},
\]
where $N$ is the number of road segments and
$\tilde{y}_i$ and $y_i$ denote an estimate and actual future observation of travel speed on road segment $i$, respectively.
We ran each experiment five times and selected the median among the five lowest validation errors.
We reported the test error from the epoch with the median validation MRE.
The report error was re-written in the table with mean absolute percentage error (MAPE) for accessability purposes;
MAPE is defined by multiplying MRE by $100$.

\bibliography{reference,aaai24}